\documentclass[lettersize,journal]{IEEEtran}
\usepackage{amsmath,amsfonts}
\usepackage{algpseudocode}
\usepackage{algorithm}
\usepackage{array}
\usepackage[caption=false,font=normalsize,labelfont=sf,textfont=sf]{subfig}
\usepackage{textcomp}
\usepackage{stfloats}
\usepackage{url}
\usepackage{verbatim}
\usepackage{graphicx}
\usepackage{cite}
\usepackage{soul}
\usepackage{xcolor}
\usepackage{siunitx}					
\usepackage{eurosym}					

\newcommand{\quotes}[1]{``#1''}
\newcommand*\xbar[1]{%
  \hbox{%
    \vbox{%
      \hrule height 0.5pt 
      \kern0.5ex
      \hbox{%
        \kern-0.1em
        \ensuremath{#1}%
        \kern-0.1em
      }%
    }%
  }%
}

\makeatletter
\DeclareRobustCommand{\atan}{%
  \operatorname{atan}%
  \@ifnextchar2{_}{}%
}
\makeatother

\DeclareSIUnit{\pers}{pers}
\DeclareSIUnit{\EUR}{\text{\euro}}
\begin{document}

\title{Automatic Navigation Map Generation for Mobile Robots in Urban Environments}
\author{Luca Mozzarelli, Simone Specchia, Matteo Corno, Sergio Matteo Savaresi
\thanks{The authors are with the Dipartimento di Elettronica, Informazione e Bioingegneria, Politecnico di Milano, 20133 Milano, Italy.
Email: \texttt{\{luca.mozzarelli, simone.specchia, matteo.corno, sergio.savaresi\}@polimi.it}\\
}}

\maketitle

\begin{abstract}
A fundamental prerequisite for safe and efficient navigation of mobile robots is the availability of reliable navigation maps upon which trajectories can be planned.
With the increasing industrial interest in mobile robotics, especially in urban environments, the process of generating navigation maps has become of particular interest, being a labor intensive step of the deployment process.
Automating this step is challenging and becomes even more arduous when the perception capabilities are limited by cost considerations.
This paper proposes an algorithm to automatically generate navigation maps using a typical navigation-oriented sensor setup: a single top-mounted 3D LiDAR sensor.
The proposed method is designed and validated with the urban environment as the main use case: it is shown to be able to produce accurate maps featuring different terrain types, positive obstacles of different heights as well as negative obstacles.
The algorithm is applied to data collected in a typical urban environment with a wheeled inverted pendulum robot, showing its robustness against localization, perception and dynamic uncertainties.
The generated map is validated against a human-made map.
\end{abstract}

\section{Introduction}
\IEEEPARstart{T}{he} last few years have been characterized by a rising interest in mobile robotics.
Particularly noteworthy is the case of autonomous delivery vehicles, which are being considered by many logistics companies.
Indeed, mobile robots could provide an effective solution to the \quotes{last-mile problem}, which remains a bottleneck in the delivery process, as well as the major cost item \cite{european_environment_agency_first_2020,ranieri_review_2018}.

In order for delivery robots to be widely adopted, however, not only do they need to reach a high level of autonomy, but the process of deploying them must be cost effective and efficient.
The goal of the deployment process is to provide the navigation algorithms a-priori information about the working environment.
This information, in the form of maps, is necessary to complete two fundamental tasks: localization and global planning.
The localization map is a representation of the environment which can be directly compared with sensor data to estimate the robot's location.
A planning or navigation map, instead, is a model of the environment which stores whether the robot can safely traverse a given location.

One could argue that the literature is rich with examples of navigation algorithms that do not require any prior information: localization can be performed online using Simultaneous Localization and Mapping (SLAM) \cite{labbe_long-term_2018,hess_real-time_2016}, while planners that work without navigation maps have been engineered to explore the environment in autonomy \cite{ivanov_uncertainty_2019}.
In both cases, however, having the information available makes a significative difference in the performance of the respective modules:
precomputing the localization map allows the SLAM system to optimize it with the whole dataset available, resulting in a more consistent and accurate localization.
Having a navigation map available is even more important, since the planner can find the shortest path to the destination directly.
Just like a human, a robotic vehicle is more efficient in navigating an environment if it has knowledge of its structure.
It is clear, then, that in a commercial delivery operation, where the goal is efficiency and profit, prior information is of fundamental importance.

A typical deployment process starts with a data acquisition phase, in which the robot explores the area of interest either autonomously or, more commonly, teleoperated by a human.
The acquired data can then be processed to build the localization and navigation maps.
As previously mentioned, the localization map is built by a SLAM algorithm, for which many options have been developed in the literature.
The output of the SLAM algorithm - the localization map - is typically obtained by aggregating and integrating raw sensor measurements.
The process of building a navigation map, instead, is not limited to merging sensor data, but can be classified as a combined mapping-perception problem.
Indeed, sensor data needs to be interpreted to mark locations as safe for traversal or as dangerous.
The perception task is particularly challenging in a urban environment, where the areas to be marked as untraversable can have a wide range of characteristics: the vehicle must detect small steps, potholes, steep ramps.
It also has to be able to mark on the map negative obstacles, like the end of sidewalks, as well as distinguishing smooth surfaces like asphalt, cement and gravel from rougher ones like grass and dirt surfaces in parks and squares.

This paper proposes an algorithm capable of automatically generating a complete navigation map from raw sensor data, including all obstacles a robot could encounter in a urban environment.
Targeting small, sidewalk-travelling robots renders the perception task quite critical, since small obstacles and minor roughness can damage them or their payload.
Furthermore, contrary to most published works, we target the usage of a single sensor for both navigability assessment and autonomous navigation, removing the need for a dedicated mapping vehicle with costly additional sensors.

The remainder of this article is structured as follows: Section \ref{sec:related_work} analyzes the related literature and justifies the sensor choice.
Section \ref{sec:experimental_setup} describes the robot used as a testbed for the proposed algorithm.
Section \ref{sec:algorithm_design} explains the algorithm itself, while Section \ref{sec:experimental_validation} presents experimental validation results.

\section{Related work}
\label{sec:related_work}

In the literature, a variety of approaches have been proposed to analyze the navigability of terrain around a robotic vehicle.
A broad classification can be arranged based on the exploited sensors: solutions using monocular, stereo or time of flight (ToF) cameras, as well as 2D laser scanners and 3D LiDARs have been proposed.

Since monocular cameras do not provide any geometric information, the navigability estimation must be based on the terrain visual characteristics.
This peculiarity constrains the type of algorithmic framework employed to be almost exclusively in the data-driven domain.
In this category falls \cite{mei_scene-adaptive_2018}, which detects unpaved road edges defining a joint classification-segmentation problem: the image is segmented in road/not-road classes using a Gaussian Mixture Model.
The specific model is selected by classifying the road type in a Markov Random Field framework with the goal of adding robustness with respect to different terrain types and illumination conditions.
Also \cite{mayuku_self-supervised_2021} uses a camera to cluster similar terrain types and assess the navigability of ground regions in front of the robot.
The training data is self-generated when the robot traverses an imaged region by taking into account accelerometer data.
The main disadvantage in using a monocular camera for terrain classification is the lack of geometric information: since navigability is ultimately defined by geometric properties, a data-driven approach with big training datasets or a complex self-training mechanism are required to correlate visual properties to geometric navigability.
This makes these approaches hard to scale to multiple terrain types.
Finally, the lack of depth information renders the navigation task challenging, if not impossible, making the objective of a single sensor setup unfeasible.

Stereo and time-of-flight (ToF) cameras can complement the visual information with geometric data.
This additional data allows more flexibility in the choice of the algorithmic framework.
Virtually all the approaches begin with simple geometric features extractions (elevation, roughness, surface normal vectors...) but, while some works achieve satisfactory results with simple thresholding rules on such features, others merely use them as inputs to probabilistic models or data driven approaches.
This trend will also hold true for laser-based sensors.
In \cite{santamaria-navarro_terrain_2015}, the authors use a ToF camera to detect non-navigable locations while traversing the environment.
Their approach employs thresholding surface normals orientations and points density to detect positive and negative obstacles, falling sharply in the thresholding category.
In the data-driven one, the effectiveness of taking into account geometric information is shown in \cite{bellone_learning_2018}, where the authors compare support vector machines (SVMs) using different feature sets to classify the navigability of space by employing stereo camera point clouds.
The classifiers using surface normals are shown to perform better than their appearance-only counterparts.
The main downsides of using depth cameras are their limited range, field of view and lower accuracy compared to laser sensors.
From a mapping perspective, this means that a longer trajectory is needed to cover the same area.
It is also clear that one depth camera is not enough for navigation, as the robot would have situational awareness just within the narrow field of view of the camera.

Moving to laser-based distance sensors, the first possibility is using a so called \quotes{laser scanner}: a LiDAR sensor with a single diode-photodetector couple (\quotes{channel}).
These are generally cheaper than the 3D counterparts but produce a single line of points.
Although the sensor produces a relatively small number of points, it is well suited to analyze ground navigability, as the points are at close range, closely spaced and aligned.
This configuration is exploited by \cite{morales_autonomous_2009}, which classifies the navigability of the points using their elevation relative to a moving average along the scan line.
The authors also propose a heuristics-based outlier rejection filter, which successfully classifies leaves on the sidewalk as traversable.
A similar thresholding approach based on elevation differences, both between the robot and the point and between neighboring points, has been proposed in \cite{hata_outdoor_2009}.
However, classification results show poor performance, mainly due to inaccuracies in the point cloud registration.
Indeed, the authors improve on these results by proposing a SVM classifier that uses the described elevation differences as input features \cite{hata_terrain_2009}, moving in the domain of data-driven approaches.
Registration uncertainty is also the motivation behind the work in \cite{chen_real-time_2015}.
The approach takes into account elevation variances in a probabilistic framework both when inserting the measurements in the elevation map and when computing the navigability index.
The method is tested with different LiDAR configurations, which shows the drawbacks of using a single 2D LiDAR as a sensor: the low number of points and relatively high speed are the cause of false positives, where obstacles have been classified as traversable.
Furthermore, going back to the goal of using a single sensor, the 2D LiDAR has a clear limitation.
Indeed, it can be pointed either towards the ground to evaluate its navigability, being blind to medium to long range obstacles, or horizontally to perform obstacle detection without sensing the terrain.

To alleviate these issues some works propose to mount the 2D LiDAR on a pan/tilt unit.
This configuration allows the integration of subsequent scans in a fully three-dimensional point cloud.
An example of this approach is \cite{ferguson_autonomous_2004}, where the robot autonomously exploring a mine periodically stops, performs a full 3D scan of the environment and plans a new trajectory.
The sensor is then brought back in a horizontal configuration to provide data for localization and obstacle avoidance.
Once the point cloud is assembled, the 2D navigability map is derived using the difference in elevation within a map cell, used as a roughness proxy.
Approaches where the robot is not required to stop are also present in the literature: \cite{montemerlo_large-scale_2006} scans individual lines analyzing their slope to segment the ground.
Scans are then merged and points in the robot's height range are considered as obstacles, while negative obstacles are detected where no points are returned.
Although this method is effective, the authors comment on the unsuitability of the sensor configuration for safe autonomous navigation: even though the pan unit provided 360\textdegree~coverage the acquisition time is inevitably slow, making dynamic obstacle avoidance arduous.

To conclude, a 3D LiDAR can be used. This option has become popular in recent years, as its cost is decreasing.
This type of sensor is ideal for navigation and localization purposes: it provides high point-count range data from all around the robot.
For mapping and ground analysis purposes, however, the point clouds are quite sparse, making the task complex.
Indeed, to counteract this shortcoming, the same tilted mounting position as 2D laser scanners is often used.
This is the case for \cite{pan_gpu_2019}, in which the authors build a elevation-variance map similarly to \cite{chen_real-time_2015}, however the navigability index is computed as a mixture of slope and height deviation from neighbors.
The same configuration is used in \cite{lee_self-training_2021}, in which the authors propose a neural network exploiting geometrical features (normal direction, height differences) extracted from an elevation map.
The method is complemented with a self-training strategy to reduce the need for hand labelled data.
In \cite{pfrunder_real-time_2017} the sensor is mounted on a \quotes{nodding} device, which sinusoidally changes its pitch angle to cover a wider field of view more densely.
It is clear that this kind of solutions hinder the sensor's ability to cover the rear side of the vehicle and, indeed, other laser scanners are fitted to cover this blind spot.
Algorithmically, this approach performs a simple occupancy grid mapping in three dimensions with light post processing to remove high obstacles that do not interfere with the navigation.
Solutions where the 3D LiDAR is mounted planarly (with the rotation axis aligned with the vertical direction) do exist in the literature, but are typically limited to high channel-count sensors or to applications to bigger vehicles \cite{chen_lidar-histogram_2017,ahtiainen_normal_2017}.
The higher number of channels increases the points density significantly, while applications to bigger vehicles are less demanding in terms of the minimum obstacle size to be detected: in both cases the perception problem is made easier.

A further distinction can be made on the types of obstacles handled by the different approaches.
Notably, many works focus on road or track boundary detection only \cite{mei_scene-adaptive_2018,bellone_learning_2018,roncancio_traversability_2014,hata_outdoor_2009}, with little consideration to other types of obstacles that could be found in more unstructured environments.
While positive obstacles can can be trivial to detect, negative obstacles require careful consideration.
Table \ref{tab:mapping/intro/literature/recap_table} presents a summary of the works cited in this literature review, with the sensors used and the types of obstacles detected.
The last column reports whether the sensor employed can also be used to perform obstacle avoidance in real scenarios.
The only method that partially accomplishes this is \cite{montemerlo_large-scale_2006}.
However, the authors state that dynamic obstacle avoidance tests failed due to deficiencies in the sensor setup.
\begin{table}[]
  \caption{Overview of the methods presented in the literature review with their features: sensors employed, whether they detect positive, negative and untraversable regions, and if the setup is suitable for both terrain analysis and complete obstacle avoidance with a single sensor.}
  \label{tab:mapping/intro/literature/recap_table}
  \begin{tabular}{|l|l|l|l|l|l|}
  \hline
  \textbf{Ref.}                               & \textbf{Sensor} & \textbf{Pos. Obs.} & \textbf{Neg. Obs.} & \textbf{Trav.} & \textbf{1 Sens.}  \\ \hline
  \cite{mei_scene-adaptive_2018}              & Mono C.         &                    &                    & \checkmark     &                          \\ \hline
  \cite{mayuku_self-supervised_2021}          & Mono C.         &                    &                    & \checkmark     &                          \\ \hline
  \cite{santamaria-navarro_terrain_2015}      & ToF C., 3D L.   & \checkmark         & \checkmark         & \checkmark     &                          \\ \hline
  \cite{bellone_learning_2018}                & Stereo C.       &                    &                    & \checkmark     &                          \\ \hline
  \cite{morales_autonomous_2009}              & 2D L.           & \checkmark         &                    & \checkmark     &                          \\ \hline
  \cite{hata_outdoor_2009,hata_terrain_2009}  & 2D L.           & \checkmark         &                    & \checkmark     &                          \\ \hline
  \cite{chen_real-time_2015}                  & 2D L., 3D L.    & \checkmark         &                    & \checkmark     &                          \\ \hline
  \cite{ferguson_autonomous_2004}             &                 &                    &                    &                &                          \\ \hline
  \cite{montemerlo_large-scale_2006}          & 2D L. pan       & \checkmark         & \checkmark         & \checkmark     & $\star$                  \\ \hline
  \cite{pan_gpu_2019}                         & 2D L.           & \checkmark         &                    & \checkmark     &                          \\ \hline
  \cite{lee_self-training_2021}               & 2D L. pan       & \checkmark         &                    & \checkmark     &                          \\ \hline
  \cite{pfrunder_real-time_2017}              & 2D L. pan       & \checkmark         &                    &                &                          \\ \hline
  \textit{\textbf{Proposed}}                  & 3D L.           & \checkmark         & \checkmark         & \checkmark     & \checkmark               \\ \hline
  \end{tabular}
\end{table}

Algorithmically, this analysis reveled the importance of exploiting geometric properties: virtually all approaches exploit them whenever available, either directly \cite{montemerlo_large-scale_2006,ferguson_autonomous_2004,chen_real-time_2015} or as features to be employed in learning-based methods \cite{bellone_learning_2018,morales_autonomous_2009,lee_self-training_2021}.
Depending on the target environments, both approaches can lead to satisfactory results, with the latter having the downside of requiring significative amounts of training data or complex self-training algorithms.

To sum-up, the main contribution of this paper is the development and experimental testing of a navigability assessment algorithm which
\begin{itemize}
  \item is capable of producing navigability maps including both positive, negative and traversability properties
  \item can detect small obstacles in a urban environment, being suitable even for small and fragile robots
  \item exploits a navigation and obstacle avoidance oriented sensor setup, comprised of a single, low channel-count 3D LiDAR sensor
\end{itemize}

\section{Experimental setup}
\label{sec:experimental_setup}

\begin{figure}[!t]
  \centering
  \includegraphics[width=2.5in]{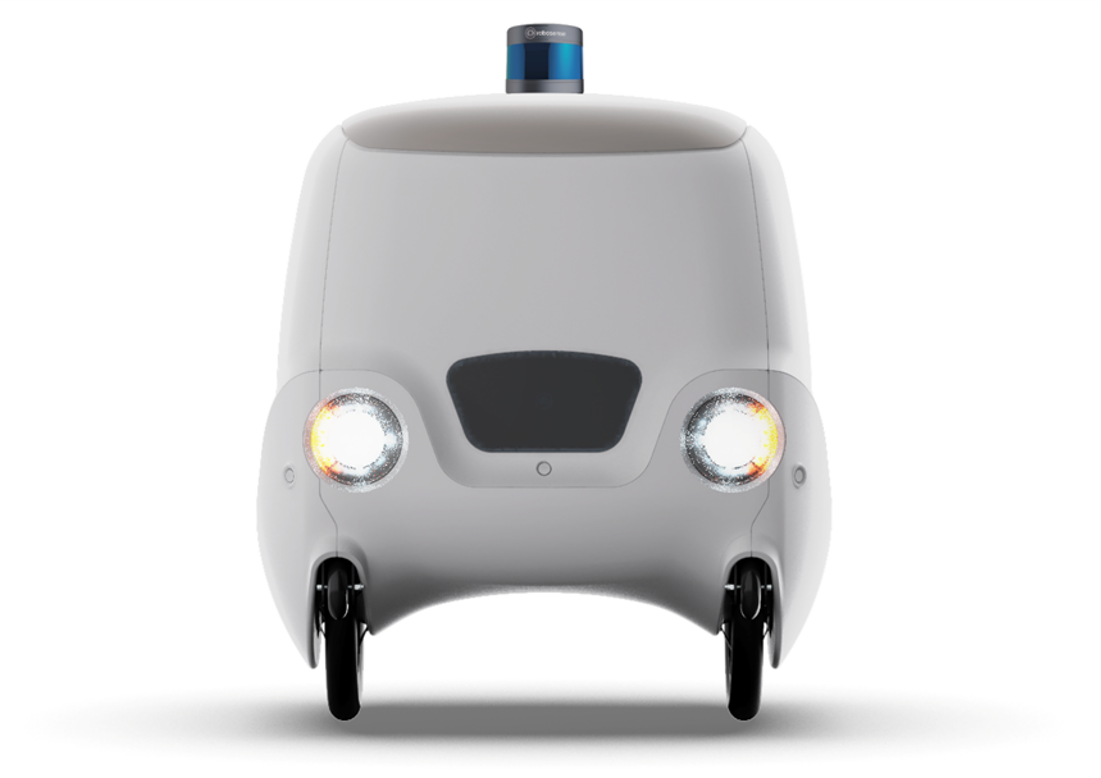}
  \caption{Yape, the robot used to test the proposed algorithm, featuring the two independent wheels and the top mounted LiDAR.}
  \label{fig:yape}
\end{figure}

The methods described herein are developed and tested on Yape, an autonomous ground drone designed for last-mile delivery (\cite{parravicini_robust_2019,parravicini_extended_2020}).
Yape (see Figure \ref{fig:yape}) is a Two Wheeled Inverted Pendulum (TWIP) robot: it features two independent driving wheels mounted to a steel chassis.
This structure makes it capable of performing agile maneuvers, but requires special care both from a vehicle dynamics perspective, which need to be stabilized, and from a perception point of view.
The pitching dynamics call for careful processing of exteroceptive sensors and introduce uncertainty when rotating the data to a non-pitching frame of reference.
These phenomena pose an additional requirement to the mapping algorithm: it should be robust to measurement and pose reconstruction errors.
The main exteroceptive sensor mounted on the vehicle is a Robosense R16 3D LiDAR.
The R16 is mounted planarly on the lid of the vehicle, as shown in Figure \ref{fig:yape}, providing 360 degree coverage of the surrounding environment and a vertical field of view of $\pm 15^\circ$ with its 16 scanning layers.

\section{Mapping Algorithm}
\label{sec:algorithm_design}

\begin{figure*}[!t]
  \centering
  \includegraphics[width=0.95\textwidth]{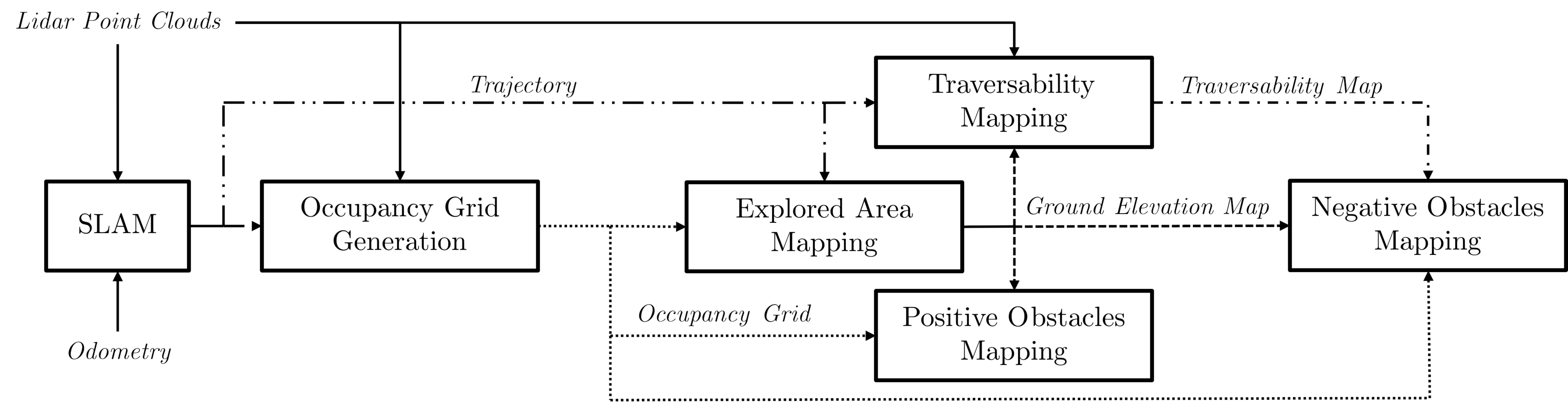}
  \caption{Block scheme of the full mapping algorithm, including the preprocessing steps.}
  \label{fig:block_scheme}
\end{figure*}
We subdivide the map generation algorithm in three main phases:
\begin{itemize}
  \item \textbf{Data acquisition}: consists in collecting sensor observations of the environment by visiting it with the robot.
  \item \textbf{Preprocessing}: the collected data points are aggregated, fusing redundant information in a computationally tractable data structure.
  \item \textbf{Navigability analysis}: raw and aggregated data are analyzed to assess the navigability characteristics of the environment.
\end{itemize}
Figure \ref{fig:block_scheme} presents a schematic view of the preprocessing and navigability analysis steps.

\subsection{Preliminary definitions}
Before delving into the algorithmic details, a definition of the employed data structures is provided.

The input on which the algorithm is based are LiDAR point clouds.
A point cloud $\mathcal P$ represents the environment as a collection of $N$ three-dimensional points $p_i = \begin{bmatrix} x_i & y_i & z_i \end{bmatrix}^T$
\begin{equation}
  \mathcal P = \left\{ p_1,  \cdots p_i, \cdots p_N \right\} \text{.}
\end{equation}

The output, instead, is a two-dimensional navigation map.
A 2D discrete map $\mathcal M$ is a grid representation of the environment:
\begin{equation}
  \mathcal M = \left\{ m_{i,j} \vert i = 1 \dots I, j = 1 \dots J \right\}
  \label{eq:2d_map}
\end{equation} 
where each cell $m_{i,j}$ represents a square portion of space of size $l \times l$, being $l$ the resolution of the map.
The overall map will then represent a space of size $I\cdot l \times J \cdot l$.
Each cell can contain different sets of information: occupancy grids \cite{moravec_high_1985} store the probability $o$ that a given cell is occupied by an obstacle 
\begin{equation}
  m_{i,j}^{OG} = \langle o \rangle
  \label{eq:occupancy_grid_cell}
\end{equation}
elevation maps store statistics about the $z$ coordinate of the points falling within the cell
\begin{equation}
  m_{i,j}^{E} = \langle z^{avg}, z^{min}, z^{max}, z^{var} \rangle
  \label{eq:elevation_map_cell}
\end{equation} 
while navigability maps can store a navigability index or, in our case, a Boolean value indicating if the cell is traversable
\begin{equation}
  m_{i,j} = \langle t \rangle \text{.}
\end{equation} 
Maps storing a single value per cell can be equivalently viewed as matrices or images, so that computer vision techniques and morphological operations can be easily applied.
Multiple maps, capturing different navigability aspects can be fused by applying a logical \quotes{and} operator.

Occasionally, the three-dimensional counterpart of the discrete map described above could be useful, especially in the case of 3D occupancy grids.
The concept of 3D occupancy grids is the same as their two-dimensional counterparts described by \eqref{eq:2d_map} and \eqref{eq:occupancy_grid_cell}.
In the 3D case, however, the cell $m_{i,j,k}$ will have cubic shape instead of being a square.

\subsection{Data acquisition}

The data necessary to the mapping algorithm is acquired while teleoperating the robot manually.
LiDAR point clouds, as well as odometric data, are recorded so that the mapping process can run offline with the complete data set.
In order to build a coherent map, the pose of the robot at the time of each LiDAR acquisition is estimated employing the Cartographer SLAM algorithm \cite{hess_real-time_2016}.

\subsection{Preprocessing}
\label{sec:preprocessing}
With the range returns and robot pose data it is possible to merge all the point clouds in a global representation of the environment.
Instead of simply merging the single point clouds and keeping the individual points, we chose to represent the environment through a three-dimensional occupancy grid.
The occupancy value $o$ is recursively updated applying the inverse sensor model \cite{elfes_using_1989} to all recorded point clouds.
This representation has two advantages over a dense point cloud: firstly, the recursive and probabilistic update allows the removal of moving objects, which would otherwise be incorrectly inserted in the map.
Secondly, it reduces the number of points and thus the computational load of the subsequent steps.
The occupancy grid is implemented as an octree \cite{hornung_octomap_2013} in order to reduce the memory requirements and raytracing times for large maps.

\subsection{Navigability analysis}
\label{sec:nav_analysis}
The characteristics of Yape make it quite sensitive to the presence of rough or unpaved terrains. These must therefore be properly considered for generating a suitable
trajectory for the robot. Consequently, the navigation map must contain all the information necessary to the proper functioning of the planning module.
In particular, the map must signal the presence of four main types of hazard:
\begin{enumerate}
  \item{ \textbf{Positive obstacles} are objects that protrude upward from the ground and can cause collisions with the robot: poles, walls, stairs, parked cars, etc.}
  \item{ \textbf{Negative obstacles} are downward jumps in the support terrain, representing a falling hazard for the vehicle. Examples include sidewalk curbs, steps (when seen from above) or potholes.}
  \item{ \textbf{Untraversable terrain} is too rough for the robot. Examples can be grass, very rough cobblestone or big potholes.}
  \item{ \textbf{Unexplored areas} cannot be classified due to the lack of information. For this reason they have to be considered unsuitable for autonomous traversal.}
\end{enumerate}
As shown in Figure \ref{fig:block_scheme}, the proposed algorithm handles each of the hazards with an independent module, which will be detailed in the coming sections.
Each module will generate a binary, two-dimensional, discrete navigability map with the overall navigability obtained by applying a logical \quotes{and} operator between the maps generated by each submodule.
\begin{equation}
 \mathcal M = \mathcal M_{pos} \wedge \mathcal M_{neg} \wedge \mathcal M_{trav} \wedge \mathcal M_{expl}
\end{equation}
where:
\begin{itemize}
  \item $\mathcal M_{pos}$ is the map of the positive obstacles
  \item $\mathcal M_{neg}$ represents the negative obstacles
  \item $\mathcal M_{trav}$ maps the traversability of the terrain
  \item $\mathcal M_{expl}$ defines the explored area
\end{itemize}

\subsection{Explored area}%
\label{subsec:explored_area}
\begin{figure}[!t]
  \centering
  \subfloat[]{\includegraphics[width=2.5in]{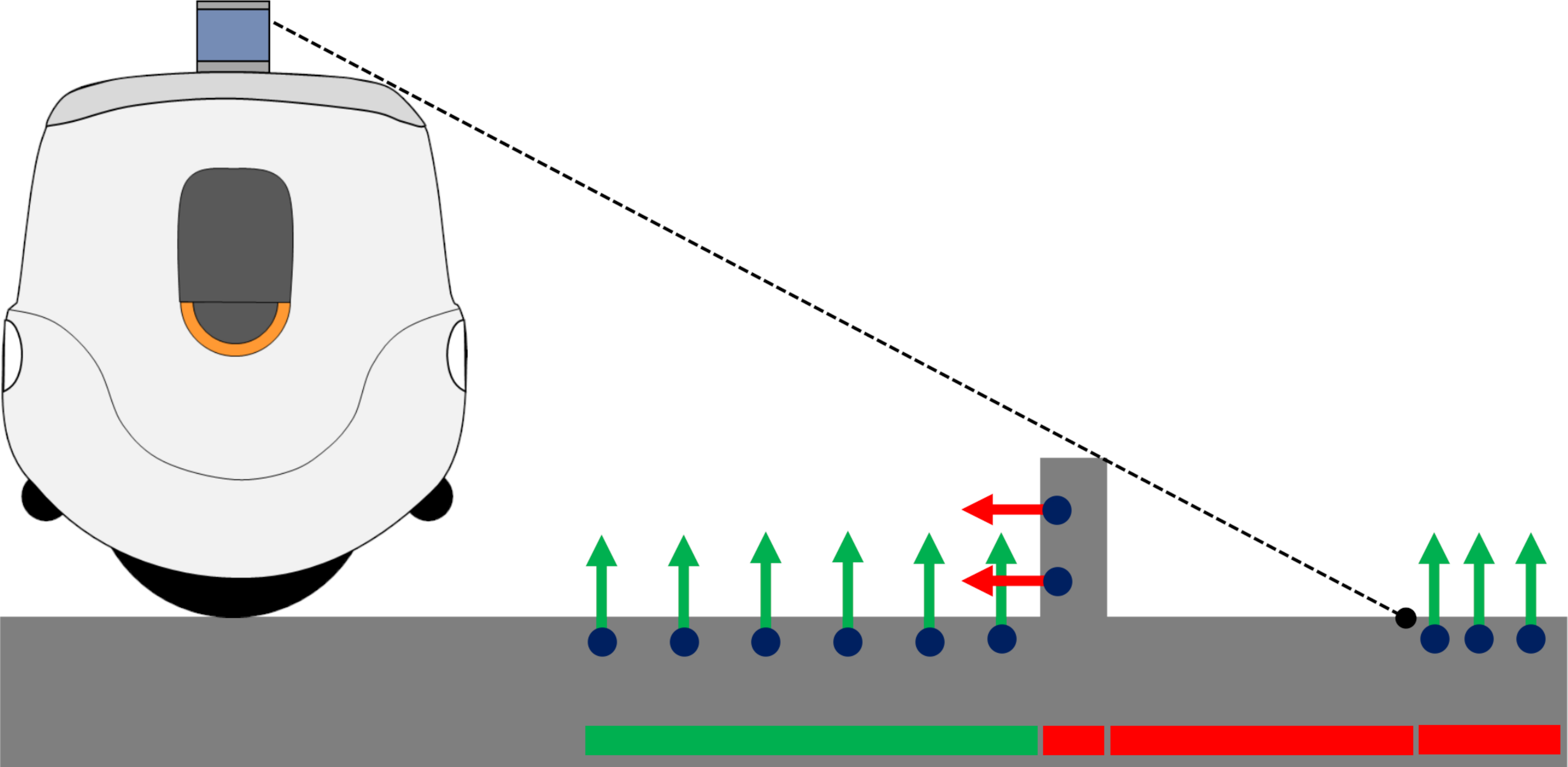}%
  \label{fig:drawing_explored_area_shade}}
  \hfil%
  \subfloat[]{\includegraphics[width=2.5in]{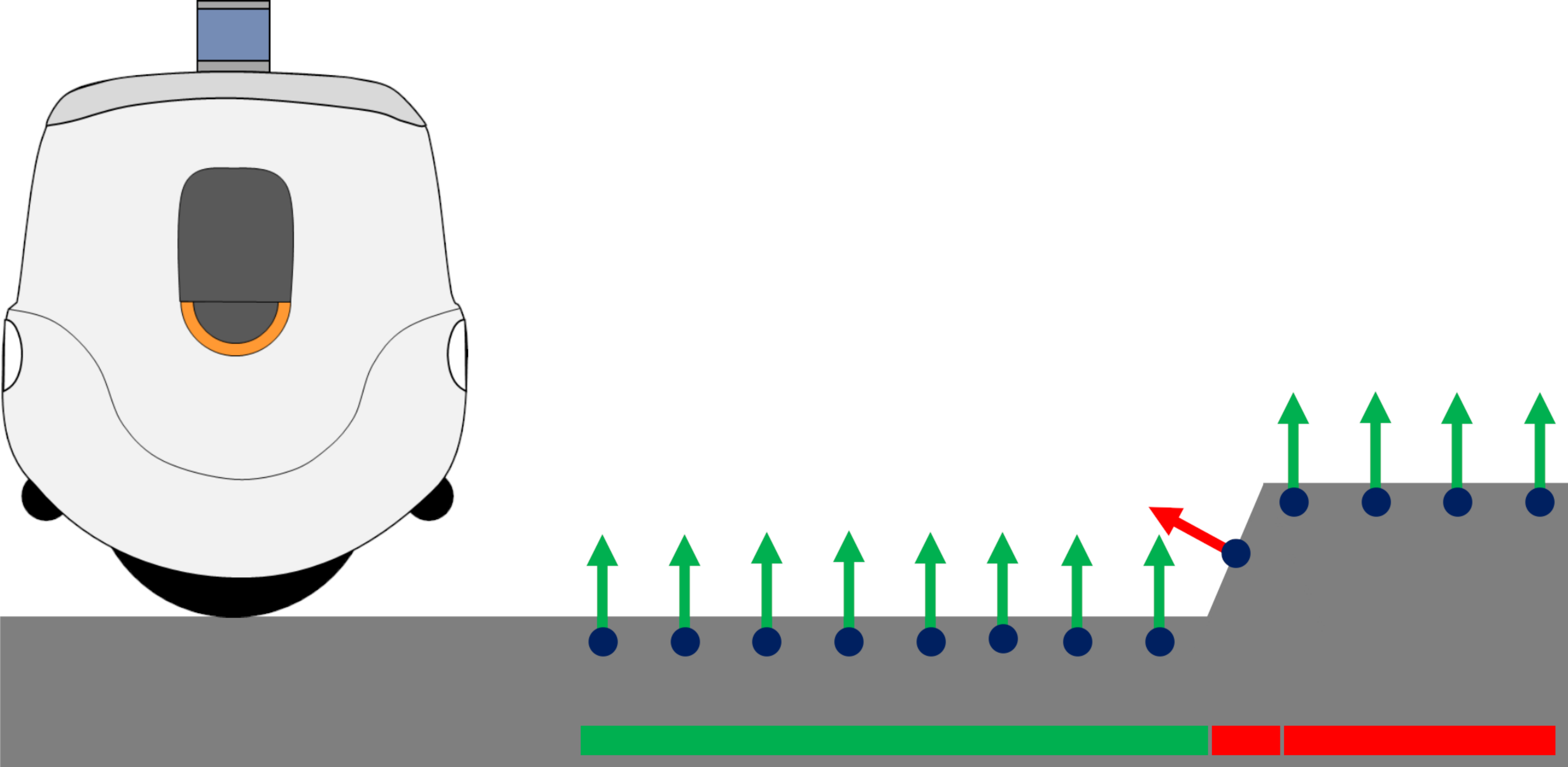}%
  \label{fig:drawing_explored_area_step}}
  \hfil%
  \caption{Schematic examples of explored area extraction. Ground regions where no measurement falls are classified as unexplored (to the right of the obstacle in \protect\subref{fig:drawing_explored_area_shade}). Ground regions which are not connected to the regions traversed by the robot are classified as unexplored (far right in \protect\subref{fig:drawing_explored_area_shade} and after the step in \protect\subref{fig:drawing_explored_area_step})}
  \label{fig:drawing_explored_area}
\end{figure}
As discussed in Section \ref{sec:nav_analysis}, information about the characteristics of the supporting terrain is needed to mark an area as safe for traversal.
For this reason, we mark an area as explored only if its ground surface can be detected.
Consequently, defining the explored area is equivalent to identifying all surfaces belonging to ground in the 3D occupancy grid obtained in Section \ref{sec:preprocessing}.
In more details, ground surfaces are identified thanks to two main assumptions:
\begin{enumerate}
  \item Ground is roughly horizontal and its slope must always be lower than the maximum slope negotiable by the robot.
  \item Reachable ground must be connected to the surface traversed by the robot during the data acquisition phase, so horizontal surfaces which cannot be reached by the robot shouldn't be considered explored.
\end{enumerate}
In order to exploit these assumptions, a lightweight representation of the environment in the form of a point cloud is firstly extracted from the 3D occupancy grid map.
Only occupied cells ($o>o^{ground}=0.5$) are selected, and the position of their center is stored (blue points in Figure \ref{fig:drawing_explored_area}).
To each point is then associated a normal vector, representing the local surface orientation (arrows in Figure \ref{fig:drawing_explored_area}).
The normal computation is performed with the Principal Component Analysis (PCA) method proposed by \cite{rusu_3d_2011} and summarized in Algorithm \ref{alg:pca_normal}.
\begin{algorithm}[H]
  \caption{PCA Normal computation \cite{rusu_3d_2011}} \label{alg:pca_normal}
  \begin{algorithmic}
    \Require Point $p$, Point Cloud $\mathcal P$, Neighbors search radius $r$
    \Ensure Normal vector $\vec n$

    \State $N=\left\{ p_1, \cdots, p_{\lvert N \rvert}\right\} \gets \textproc{PointNeighbors}(p, \mathcal{P}, r)$

    \State $ N \gets N \cup \{p\}$

    \State $\bar p \gets \frac{1}{\lvert N \rvert} \sum_{i=1}^{\lvert N \rvert} p_i$ 
    \Comment{Centroid}

    \State $C \gets \frac{1}{\lvert N \rvert} \sum_{i=1}^{\lvert N \rvert} \left(p_i - \bar p\right) \left(p_i - \bar p\right)^T$
    \Comment{Covariance matrix}

    \State $\Lambda, V \gets \textproc{eig}(C)$
    \Comment{Eigenvalues and eigenvectors}

    \State $i \gets \text{argmin}_i(\lambda_i \in \Lambda)$
    \Comment{Minimum eigenvalue}

    \State $\vec n \gets v_i \in V$
    \Comment{Eigenvector with minimum eigenvalue}

    \State \Return $\vec n$
  \end{algorithmic}
\end{algorithm}
The slope of the surface can then be evaluated by computing the angle between the normal vector and the vertical versor $\vec z=\begin{bmatrix} 0 & 0 & 1 \end{bmatrix}^T$:
\begin{equation}
  \alpha' = \arccos\left(\frac{\vec n \cdot \vec z}{\lvert n \rvert}\right).
  \label{eq:slope_0_180}
\end{equation}
Two observations are in order regarding \eqref{eq:slope_0_180}.
The obtained value of $\alpha'$ is unsigned: it will take values between $0$ and $\pi$ irrespective of the rotation sign between $\vec n$ and $\vec z$.
This does not constitute an issue in our application, since we are interested in identifying both climbs and descents.
Secondly, algorithm \ref{alg:pca_normal} does not provide consistent normal direction (i.e. out from the surface or into the surface).
Under the previous assumption of making no distinction between climbs and descents, we cast all angles back in the range $\left[0,  \pi/2\right]$ with \eqref{eq:slope}.
\begin{equation}
  \alpha = \begin{cases}
    \alpha' & \text{if } \alpha' \leq \frac \pi 2 \\
    \pi - \alpha' & \text{otherwise} \\
  \end{cases}
  \label{eq:slope}
\end{equation}
All points with angle $\alpha$ lower than a threshold $\alpha^{ground} = 30^\circ$ are candidate ground points.
This method, however, identifies all horizontal surfaces, including ceilings and other ground surfaces not reachable by the robot (think of an elevated sidewalk without a connecting ramp).
To discard this kind of false negatives the second characteristic of \quotes{valid} ground surfaces is exploited: we aim to discard all the surfaces that would not be reachable by the robot starting from the mapping trajectory.
With this goal an Euclidian clustering algorithm, as implemented in the Point Cloud Library \cite{rusu_3d_2011}, is applied to the candidate ground points.
Note that the algorithm works on all three dimensions, so that close-by surfaces at different elevation can be separated.
Each returned cluster%
is then checked against the trajectory: if the trajectory intersects the cluster, the latter is identified as ground.
The intersection is checked by comparing the horizontal and vertical distance between the trajectory points and the cluster points with the thresholds $d_{xy}^{ground}$ and $d_{z}^{ground}$, respectively.
To obtain a 2D binary map, the points are projected on the $xy$ plane, marking a cell as traversable if at least one ground point falls within its limits.
Finally, a morphological closure is applied to the obtained map to remove noisy points and filling the gaps caused by incorrect sensor pose estimation.

The identified ground is also exploited to build a second output from this block: an elevation map.
The elevation map, as defined in equations \eqref{eq:2d_map} and \eqref{eq:elevation_map_cell} will define a ground elevation reference exploited by the other algorithm blocks.

\subsection{Positive Obstacles}
\label{subsec:positive_obstacles}
\begin{figure}[!t]
  \centering
  \includegraphics[width=2.5in]{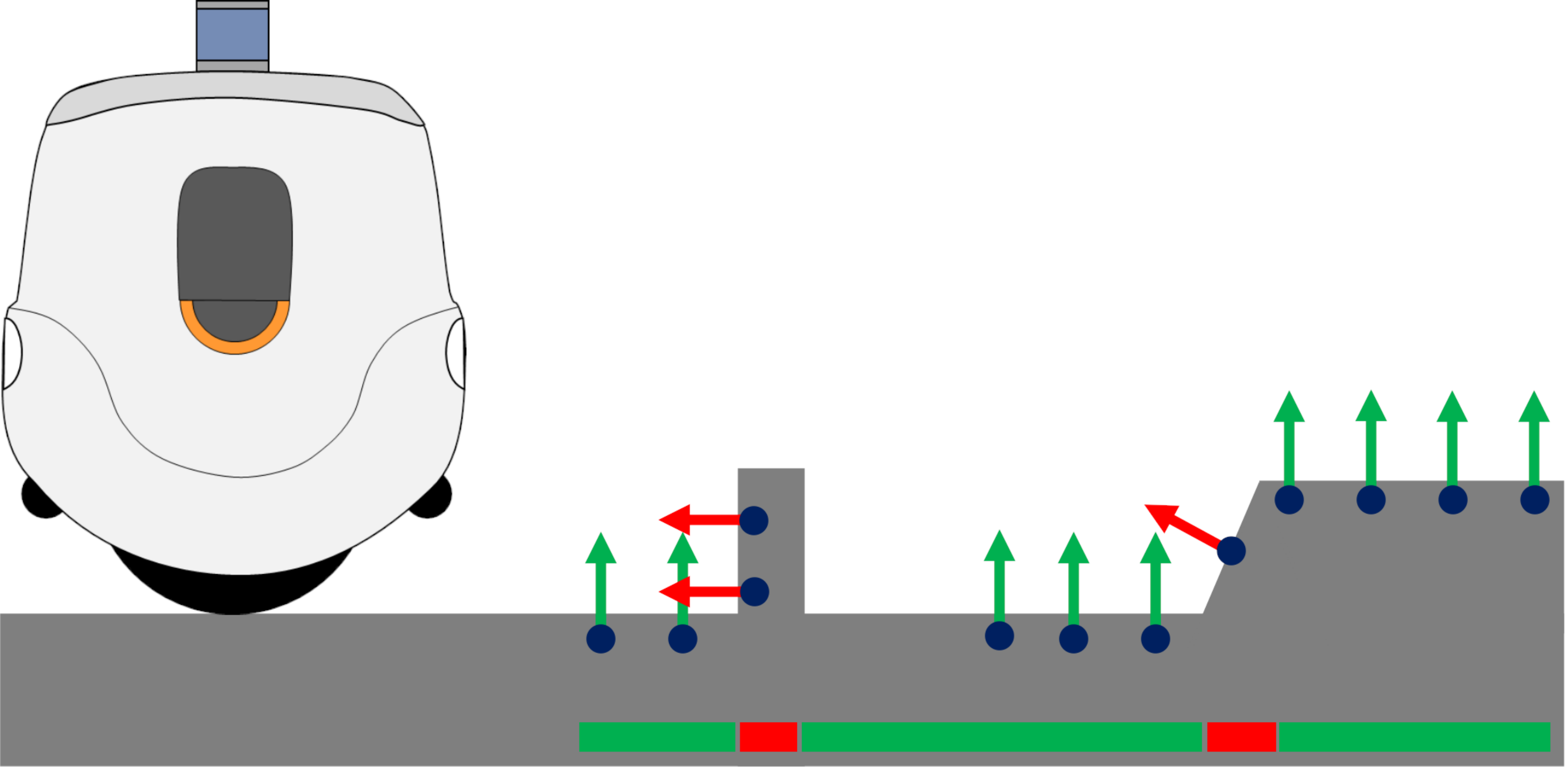}%
  \caption{Schematic examples of positive obstacles extraction. Areas with vertical or too steep slopes are classified as obstacles (red underline).}
  \label{fig:drawing_positive_obstacles}
\end{figure}
Positive obstacles are detected in a similar but complementary way with respect to ground points: once the 3D occupancy grid is converted to a point cloud, surface normals are computed and all points with a normal angle larger than a threshold $\alpha^{obs}=15^\circ$ are candidate obstacles (see Figure \ref{fig:drawing_positive_obstacles}).
Note that the occupancy threshold is higher for obstacles than it is for ground points in order to reduce false negatives.
Next, a height filter is applied to prevent the inclusion of obstacles which are too high to cause any real danger to the robot (consider tree branches protruding over the sidewalk as an example).
In particular, for each point, the corresponding ground elevation $z_{ground}$ is extracted from the ground elevation map described in Section \ref{subsec:explored_area}.
Then the point is discarded if $z_{point} - z_{ground} > z_{th}$, with $z_{th}$ set to the robot height plus a safety margin. 
Finally, noise is removed by applying again the Euclidian clustering algorithm \cite{rusu_3d_2011} and discarding clusters with fewer than $N^{obstacles}_{min~size}$ points. 
The positive obstacles map %
is obtained by marking as not traversable the cells in which at least one obstacle points falls.

\subsection{Untraversable Terrain}
\label{subsec:traversability}
\begin{figure*}[!t]
  \centering
  \subfloat[]{\includegraphics[height=1.5in]{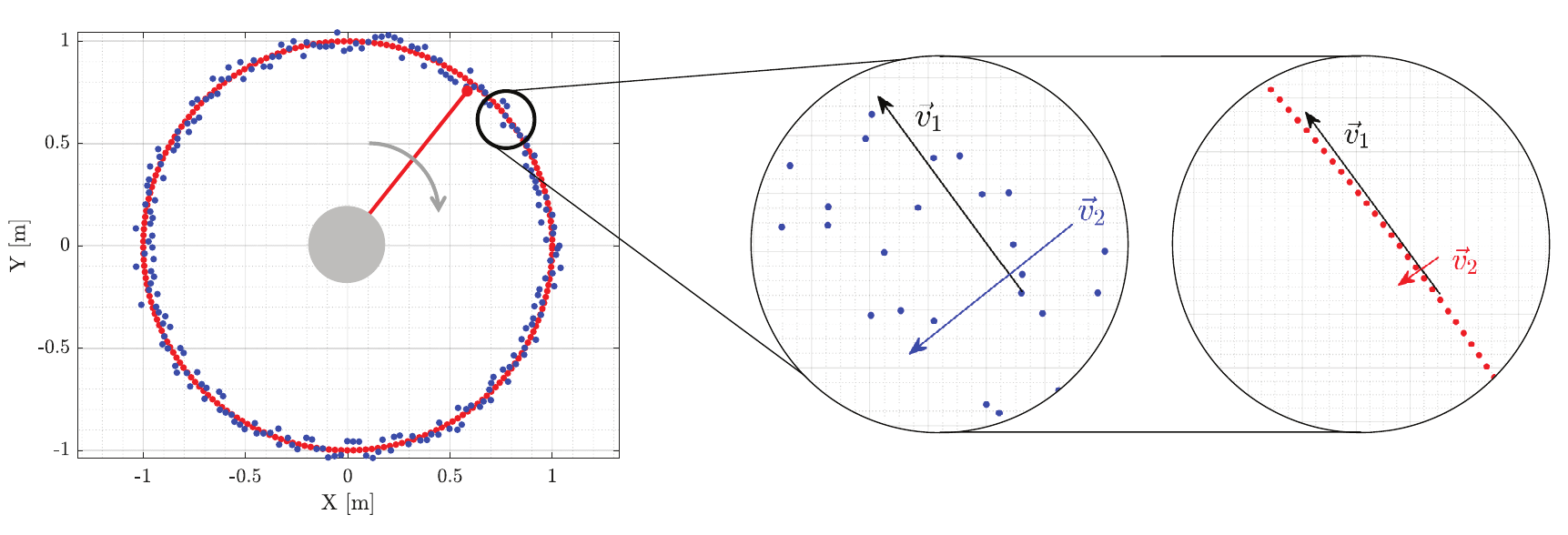}%
  \label{fig:lidar_roughness}}%
  \hfill%
  \subfloat[]{\includegraphics[height=1.7in]{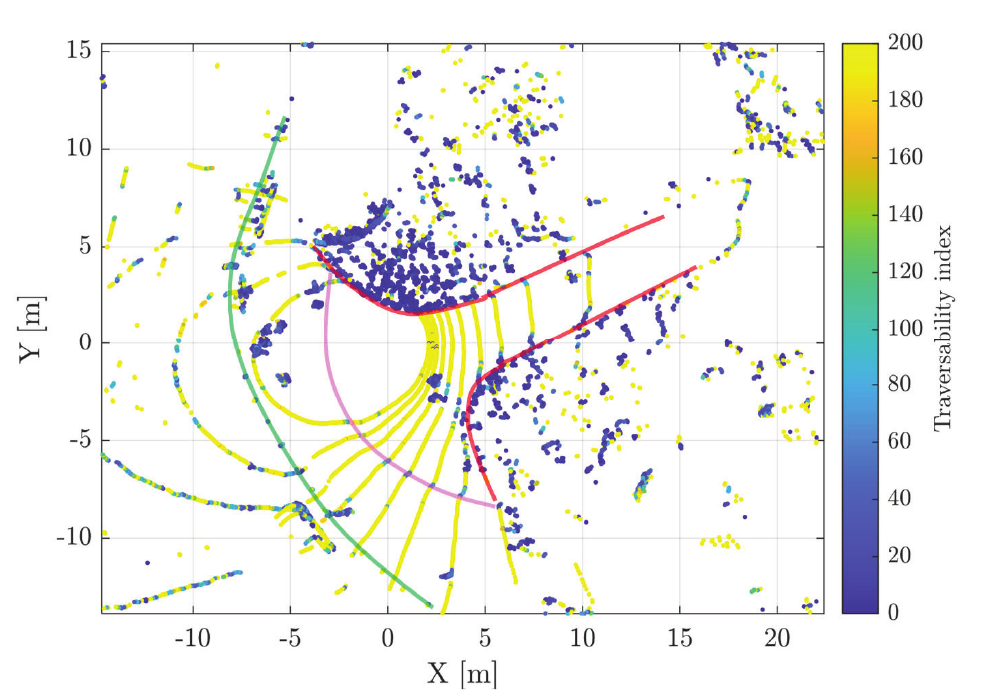}%
  \label{fig:piazza_leonardo_traversability_index_single_point_cloud}}
  \caption{Simulated LiDAR \quotes{ring} on a planar surface illustrating the effect of surface roughness \protect\subref{fig:lidar_roughness} and a point cloud color coded with the traversability index value \protect\subref{fig:piazza_leonardo_traversability_index_single_point_cloud}. Red annotated lines show the boundary between gravel and grass, pink line between gravel and asphalt and green between sidewalk and cobblestone road.}
\end{figure*}

A traversability analysis is performed to evaluate the ground surface roughness, intended as the presence of
irregularities and small height variations in the supporting terrain, that can cause excessive stress on the drone's components. 
Since these typically consist of sub-centimeter variations, it is clearly not possible to conduct the analysis starting from the 3D occupancy grid: the discretized representation does not retain this information.
Rather, we perform the analysis at the point cloud level.
The idea behind the proposed traversability index is to exploit the effect of terrain irregularities on the shape of LiDAR \quotes{rings}.
The concept is illustrated in Figure \ref{fig:lidar_roughness}.
Assuming a planar ground, each LiDAR channel draws a curve on the plane.
By considering a small neighborhood on the curve, points will appear distributed along a line.
If the surface is perfectly smooth the fitting to the line model will be ideal; as the roughness increases, some of the points will be detected at closer range (those with elevation higher than the flat ground), while others (belonging to dips in the ground) will appear further away.
The effect on the line (when viewed from above) will be similar to an added noise in the radial direction: the points will still have a major distribution axis, that is a linear shape, but the deviation along the perpendicular direction will increase.
By measuring this deviation, we can have a proxy of the surface roughness.

In practice, point clouds are first filtered in height, with the aim of discarding points that cannot belong to the ground.
The process is similar but dual with respect to the height filter in Section \ref{subsec:positive_obstacles}.
Then, the points are projected on the $xy$ plane by setting their $z$ value to $0$.
The 2D version of Algorithm \ref{alg:pca_normal} can be applied to compute the described proxy: the roughness index $t$ of point $p$ will be the ratio between the maximum and minimum eigenvalues.
\begin{equation}
  t(p) = \frac{\max\left(\lambda_1, \lambda_2\right)}{\min\left(\lambda_1, \lambda_2\right)}
\end{equation}
Figure \ref{fig:piazza_leonardo_traversability_index_single_point_cloud} shows a point cloud color coded with the traversability index.
Once each point is tagged with its traversability index, a map is created by down projecting the points and averaging the traversability values of points falling in the same cell.
Note that the resulting map is not binary, but it can be thresholded.
The simplest option would be to consider traversable all cells with a traversability index above a fixed threshold $t_{traversable}$.
This method leads to acceptable results on most terrains, however a tuning process has to be performed on a map-by-map basis since the evaluation of the robot capabilities on different terrains is not a trivial task.
For this reason a dynamic thresholding method is proposed, in particular we assume that during the data acquisition phase, the drone only travelled along traversable terrain.
The key idea, then, is to consider safe all terrains with a similar traversability to the ones experienced by the robot during mapping.
To this end, a traversability index is associated to each pose in the robot trajectory: the robot's footprint is superimposed to the traversability index map
and the pose is marked with the average index value of the cells covered by the footprint.
The thresholding rule is then expressed by equation 
\begin{equation}
  \text{traversable}(i,j) = t(i,j) \geq \gamma \cdot t_T \land t(i,j) \geq t_{min}
  \label{eq:traversability_threshold}
\end{equation}
Where $t(i,j)$ is the traversability value of cell in position $(i,j)$, $t_T$ is the traversability index of the closest trajectory point and $\gamma=0.75$ and $t_{min}=200$ are two tuning variables.
The intuitive meaning of \eqref{eq:traversability_threshold} is that, for a point to be considered traversable, its index should be close enough ($75\%$ in our case) to the one of the trajectory that mapped it, while being always over a safety traversability value.
Similarly to the explored area map a morphological closing is applied to the resulting binary map to remove noisy points.

\subsection{Negative Obstacles}
\label{subsec:negative_obstacles}
\begin{figure}[!t]
  \centering
  \subfloat[]{\includegraphics[width=2.5in]{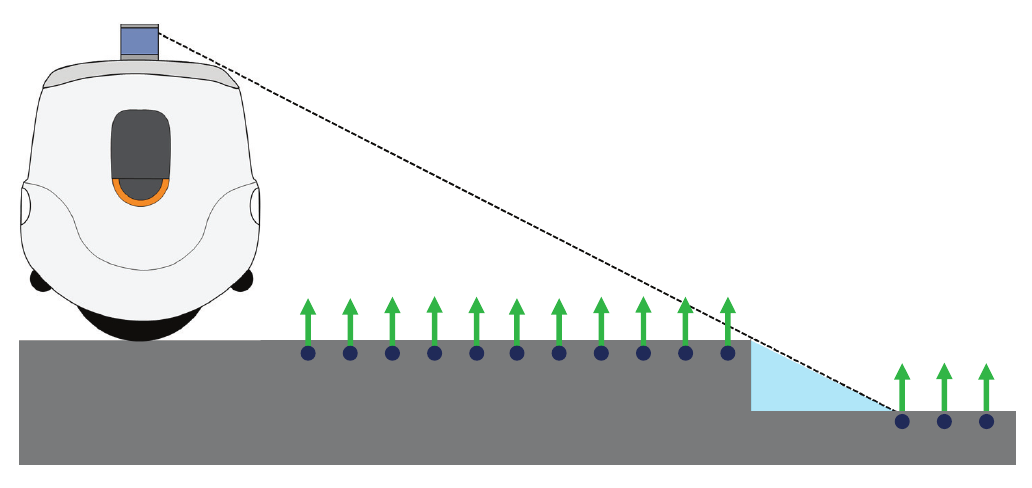}%
  \label{fig:negative_obstacles_drawing_before}}
  \hfil%
  \subfloat[]{\includegraphics[width=2.5in]{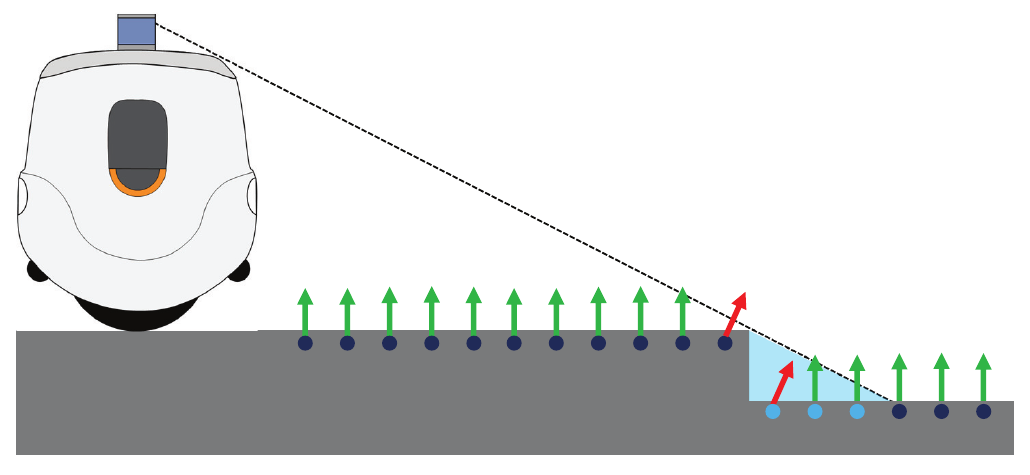}%
  \label{fig:negative_obstacles_drawing_after}}
  \hfil%
  \caption{Schematic view of a negative obstacle perception problem: without the expansion \protect\subref{fig:negative_obstacles_drawing_before}, no points are inside the shadowed area (light blue triangle), as a consequence no inclined normal is found in the centroids point cloud (dark blue dots and green arrows). After the expansion \protect\subref{fig:negative_obstacles_drawing_after}, points are added inside the shadowed area (light blue points) and inclined normals (red arrows) can be computed near the step.}
  \label{fig:negative_obstacles_drawing}
\end{figure}
Negative obstacles are particularly challenging to detect with range sensors: the vertical surface itself is not visible directly when observed from above.
Rather, the only visible effect is a shadowed area, where no measurements are returned.
Indeed, the blind area is the reason why the negative obstacles cannot be detected by the positive obstacles algorithm: within a neighborhood of the available points no height jumps occurs, so the normal will remain vertical (see figure \ref{fig:negative_obstacles_drawing_before}). 
The proposed solution is to artificially fill-in the blind spot by recursively expanding the map.
Adding points in the blind area makes the elevation jump visible to the normals computation algorithm, allowing the detection of the negative obstacle (figure \ref{fig:negative_obstacles_drawing_after}) with the algorithm already presented in Section \ref{subsec:positive_obstacles}.
The downside of expanding the map is that false obstacles could appear in all regions were the ground surface presents \quotes{holes} due to measurement noise.
Such errors are removed by an additional filtering step based on the idea that, due to the difference in height between the two surfaces connected by the negative obstacle, it will generate a small untraversable patch in the traversability layer.
While the patch typically covers the negative obstacle just partially, limited to portions of the obstacle in the vicinity of the robot, it can provide an indication of where to expect them.

Entering in more details, in order to perform the expansion we convert the 3D occupancy grid in a multiple elevation map: a 2D map in which each cell stores a vector of elevation values, each corresponding to an occupied cell in the 3D occupancy grid.
Then, all the cells with no elevation values are analyzed: for each, the minimum elevation among its neighbors is copied in the unknown cell.
The procedure is repeated $N_{expansions}$ times in order to propagate the elevation to a distance higher than the grid resolution.
Once the expansion is completed, all the elevation values of unknown cells but the minimum one are discarded and the map is converted back to a point cloud, which is then fed to the positive obstacles algorithm.
To reject the false obstacles, we apply the Euclidian clustering algorithm first.
Then, we discard all negative obstacles clusters except those that intersect a non traversable region.
Note that the negative obstacles layer does provide some redundancy in the detection of positive obstacles: some of them will be detected twice.
Of course, this is not an issue, instead it increases the probability of detecting all obstacles.

\subsection{Exploiting the mapping trajectory}
The mapping experiments provide an additional source of information: the mapping trajectory.
Given that the trajectory has been covered by the robot, we can exploit this data to reduce the number of navigable points misclassified as dangerous.
With this goal a navigability map is generated by marking as traversable all points around the trajectory, considering the robot's footprint.
The map is then merged with the partial maps using a logical \quotes{or} operator.
The merging occurs before any morphological closing operation.
Finally, the map is merged again as a final step in the algorithm.

\section{Experimental validation}
\label{sec:experimental_validation}
\begin{figure}[!t]
  \centering
  \includegraphics[width=.48\textwidth]{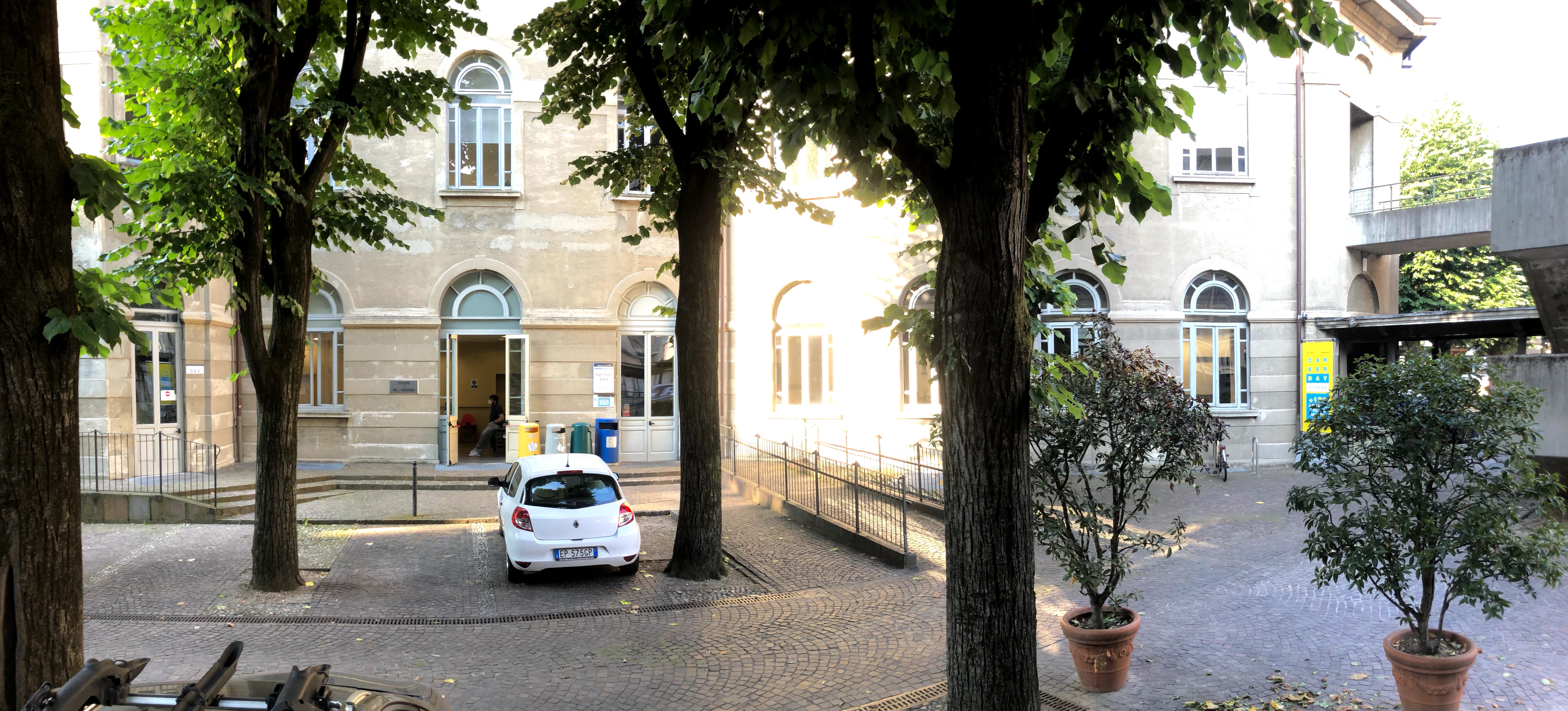}%
  \caption{Picture of the first validation map, acquired between buildings 7 and 9 of Politecnico di Milano campus.}
  \label{fig:building_9_image}
\end{figure}
\begin{figure*}[!t]
  \centering
  \subfloat[]{\includegraphics[height=1.4in]{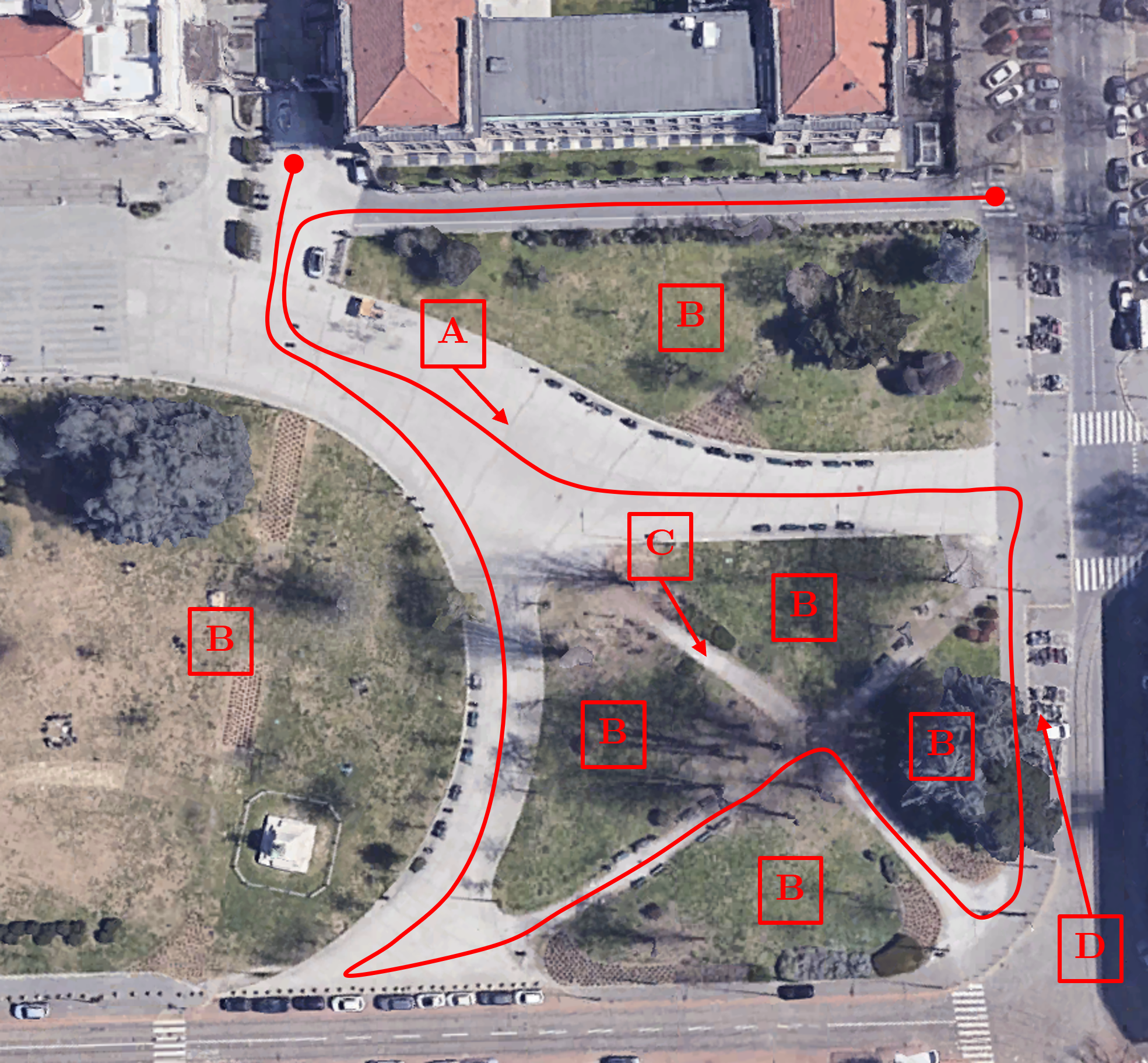}%
  \label{fig:piazza_leonardo_aerial}}
  \hfil%
  \subfloat[]{\includegraphics[height=1.4in]{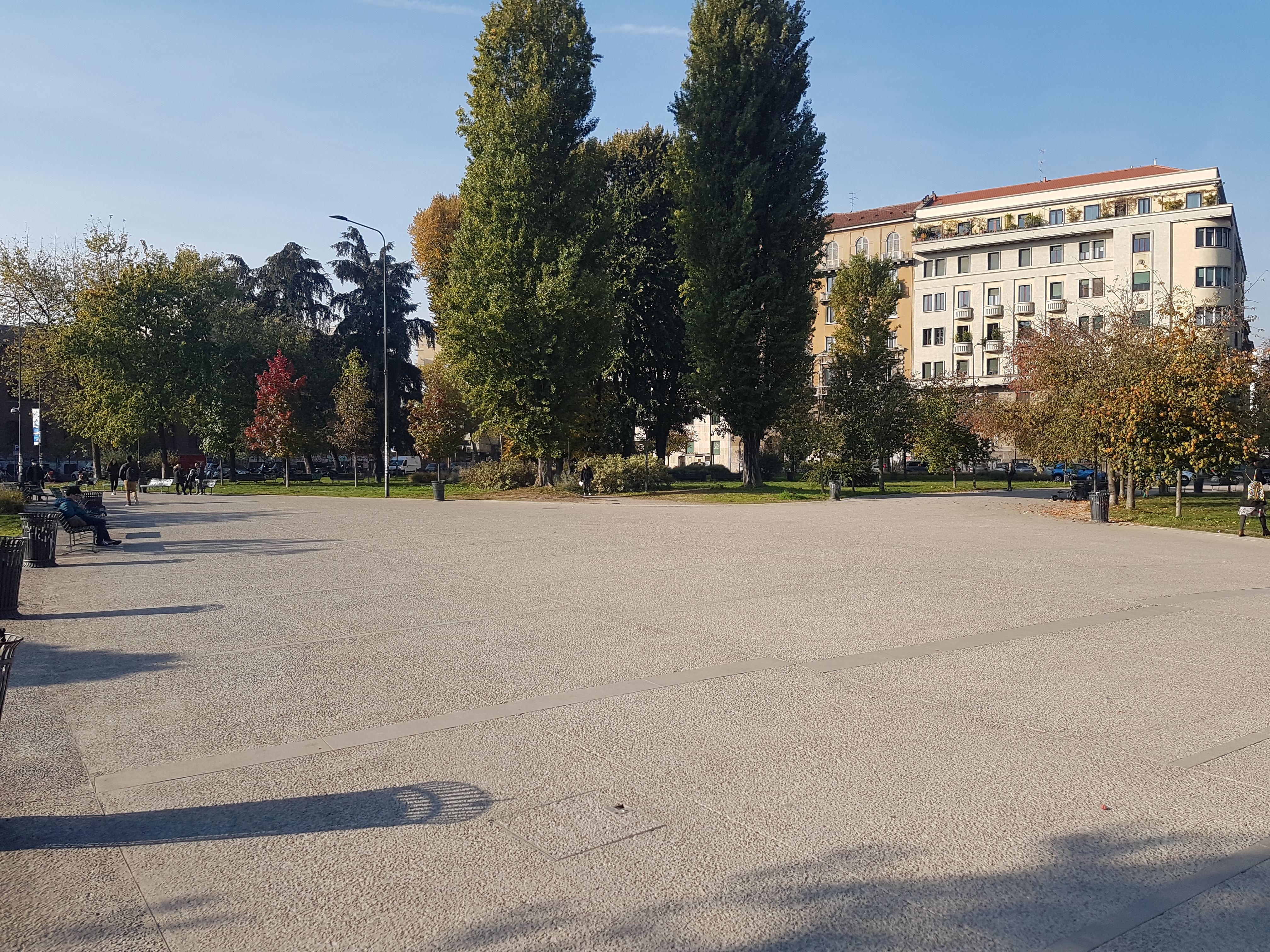}%
  \label{fig:piazza_leonardo_porous_asphalt}}
  \hfil%
  \subfloat[]{\includegraphics[height=1.4in]{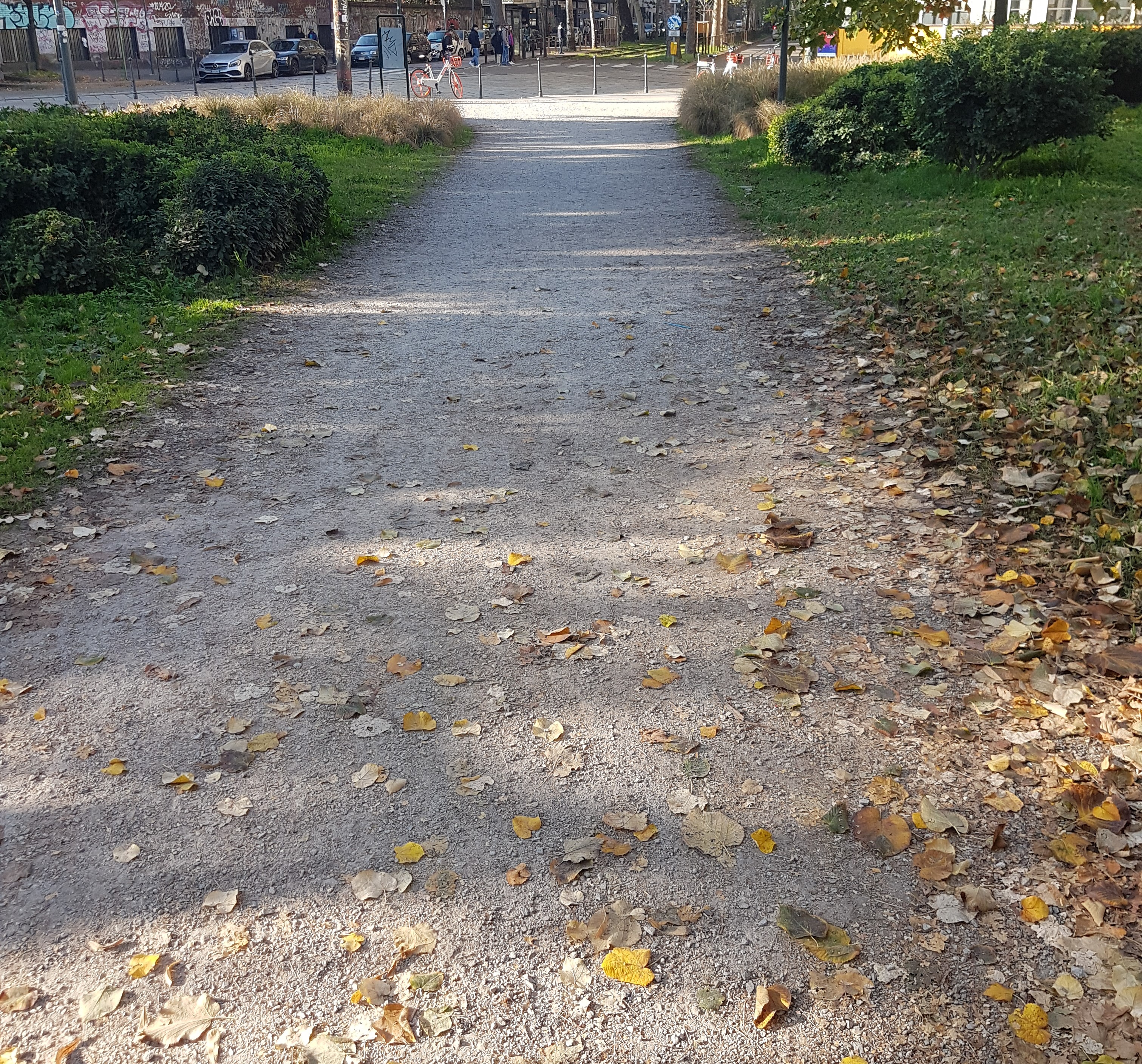}%
  \label{fig:piazza_leonardo_gravel}}
  \hfil%
  \subfloat[]{\includegraphics[height=1.4in]{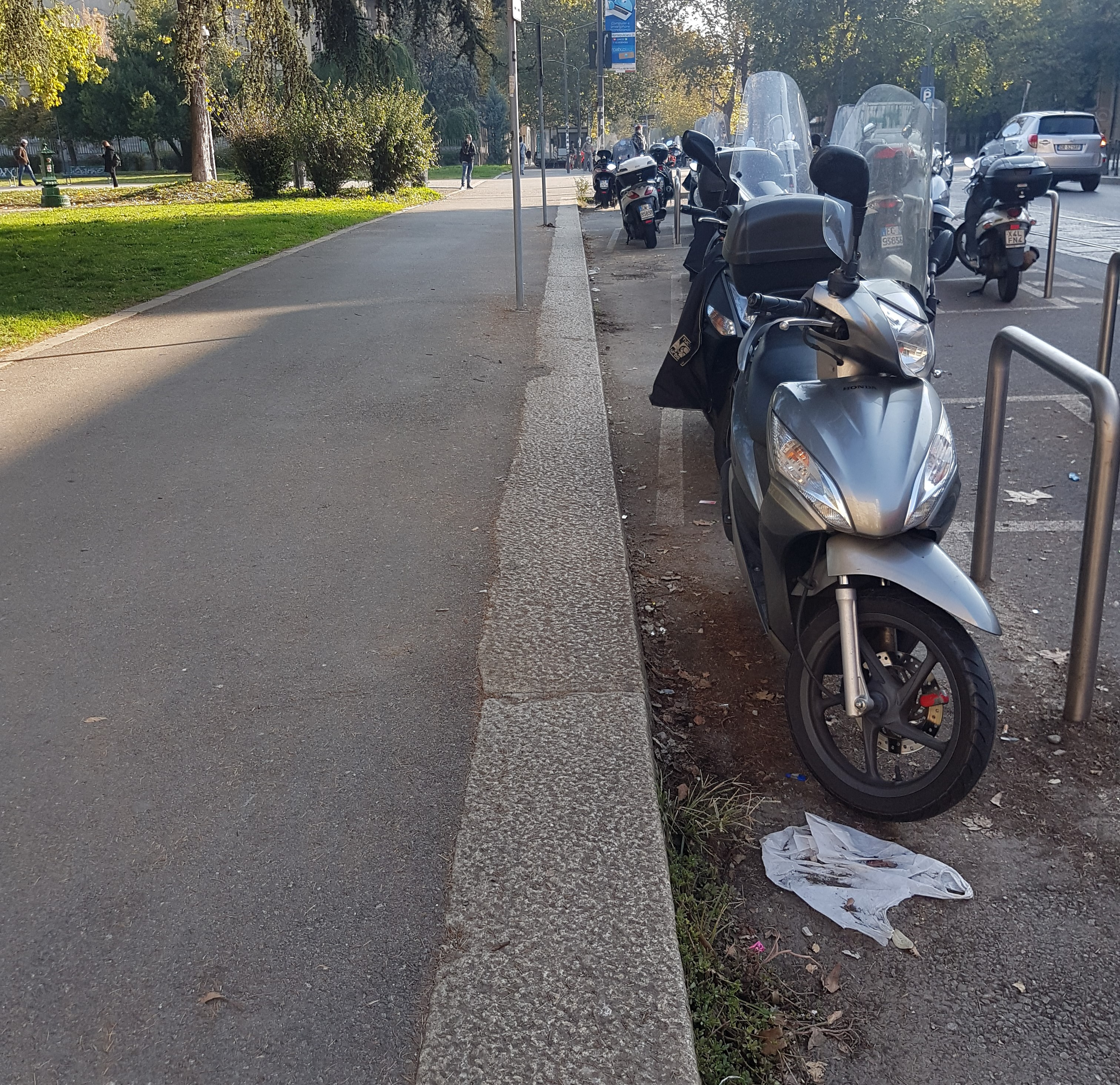}%
  \label{fig:piazza_leonardo_negative_obstacle}}
  \caption{Aerial view \protect\subref{fig:piazza_leonardo_aerial} and pictures \protect\subref{fig:piazza_leonardo_porous_asphalt},\protect\subref{fig:piazza_leonardo_gravel}, \protect\subref{fig:piazza_leonardo_negative_obstacle} of the second mapped area in Piazza Leonardo Da Vinci, Milano.}
  \label{fig:piazza_leonardo_pictures}
\end{figure*}
In this paragraph a qualitative (Section \ref{subsec:qualitative_validation}) and quantitative (Section \ref{subsec:quantitative_validation}) analysis of the experimental application of the proposed algorithm will be conducted.
Two maps will be used to illustrate the algorithm steps.
Pictures and aerial views are shown in Figures \ref{fig:building_9_image} and \ref{fig:piazza_leonardo_pictures}.
The first map, acquired in a cobblestone area between buildings 7 and 9 of Leonardo Campus, Politecnico di Milano, will aid in the illustration of the first two blocks: Explored Area extraction and positive obstacles.
The area is characterized by ground surfaces on different levels, connected by slopes steps and showing surfaces with different inclinations and obstacles of various heights.
The second map represents the southern portion of Piazza Leonardo da Vinci, in Milan and will be used to illustrate the remaining modules, as well as to perform the quantitative validation.
It presents different surface types (grass, gravel, cement, asphalt), as well as positive obstacles of different sizes (poles, litter bins, benches) and a typical negative obstacle (the sidewalk edge).

\subsection{Qualitative validation}
\label{subsec:qualitative_validation}
\begin{figure}[!t]
  \centering
  \includegraphics[width=0.48\textwidth]{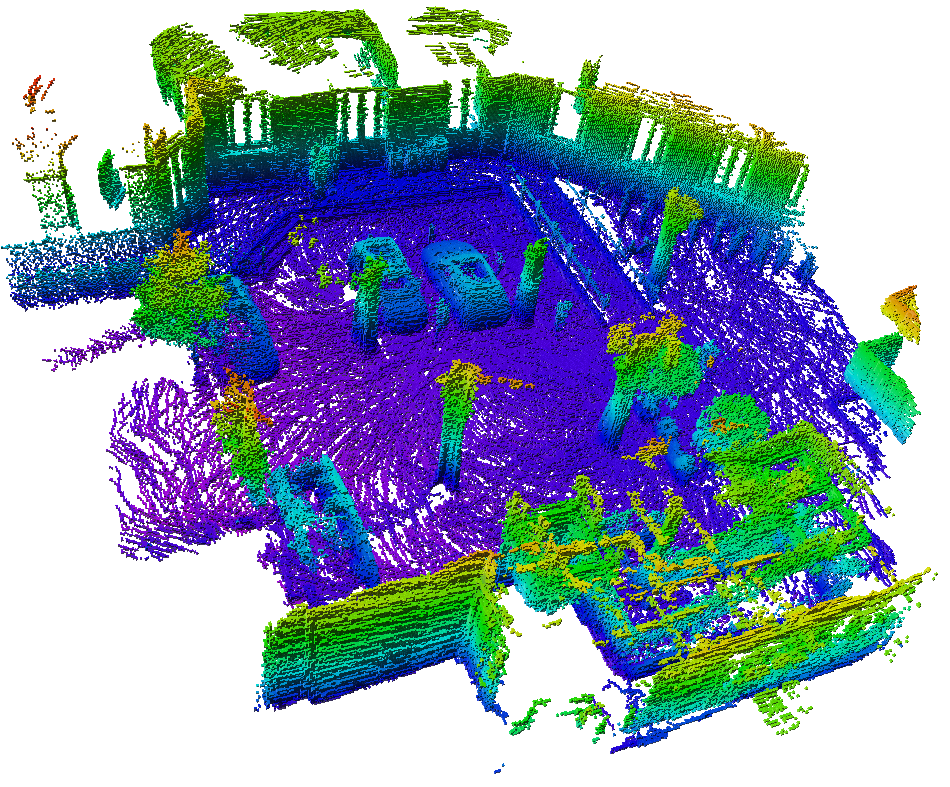}
  \caption{Three-dimensional occupancy grid of the first map. Color code represents the $z$ coordinate, only cells with occupancy probability $o>0.5$ are represented.}
  \label{fig:building_9_octree}
\end{figure}

\begin{figure*}[!t]
  \centering
  \subfloat[]{\includegraphics[width=0.254\textwidth]{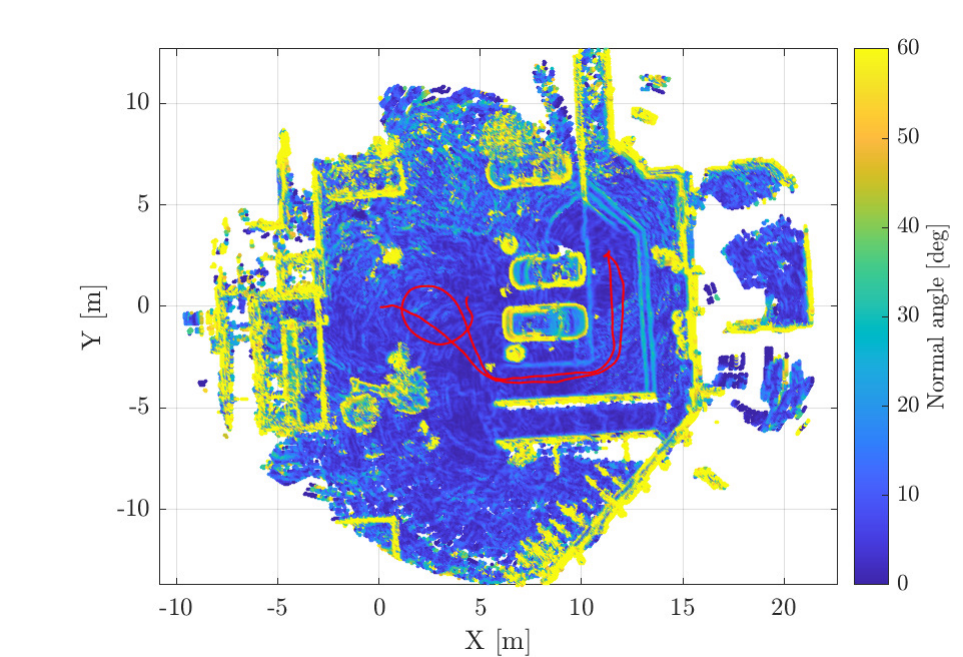}%
  \label{fig:building_9_ground_normals}}%
  \hfil%
  \subfloat[]{\includegraphics[width=0.232\textwidth]{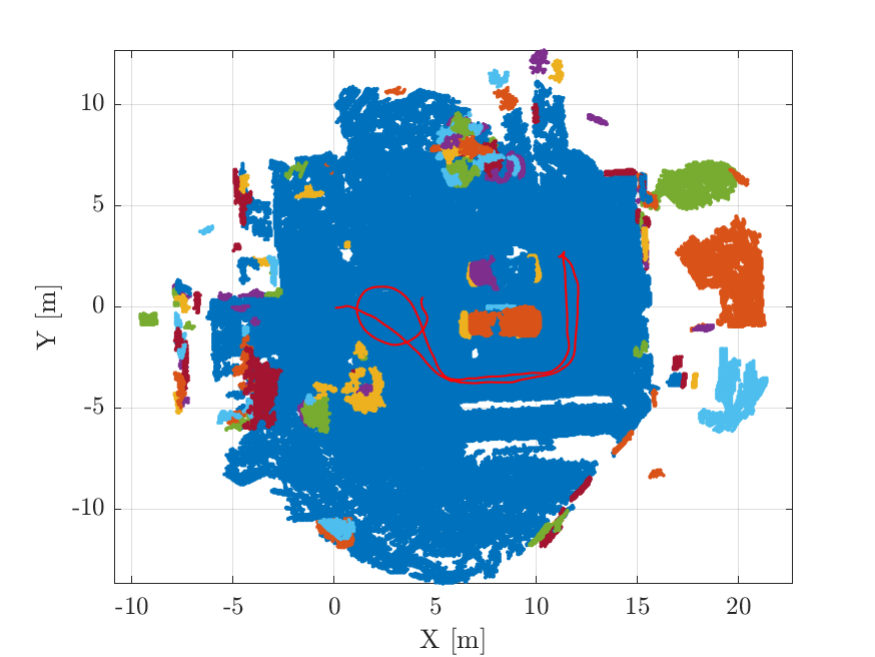}%
  \label{fig:building_9_ground_normals_flat_clustered}}%
  \hfil%
  \subfloat[]{\includegraphics[width=0.254\textwidth]{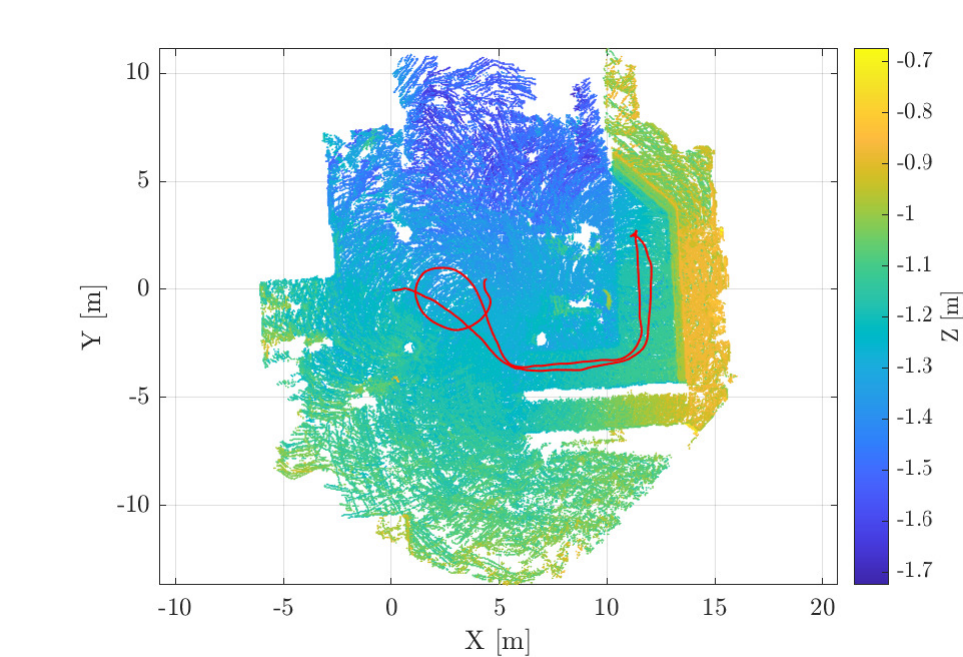}%
  \label{fig:building_9_ground_elevation}}
  \hfil
  \subfloat[]{\includegraphics[width=0.232\textwidth]{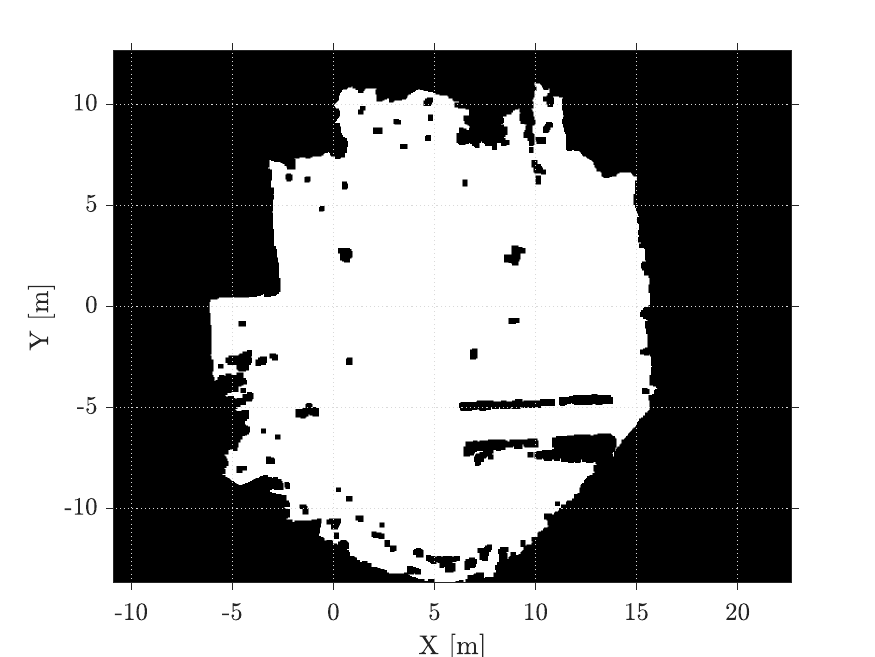}%
  \label{fig:building_9_ground_map_inflated}}
  \hfil%
  \caption{Intermediate steps of the explored area pipeline. Centroids of the occupied cells color coded with the normal angle \protect\subref{fig:building_9_ground_normals}, points with low normal angle after the Euclidian clustering step \protect\subref{fig:building_9_ground_normals_flat_clustered}, ground cluster with elevation information \protect\subref{fig:building_9_ground_elevation} and the final explored area map after the closure \protect\subref{fig:building_9_ground_map_inflated}. The red line represents the mapping trajectory.}
  \label{fig:building_9_ground}
\end{figure*}

\begin{figure*}[!t]
  \centering
  \subfloat[]{\includegraphics[width=0.249\textwidth]{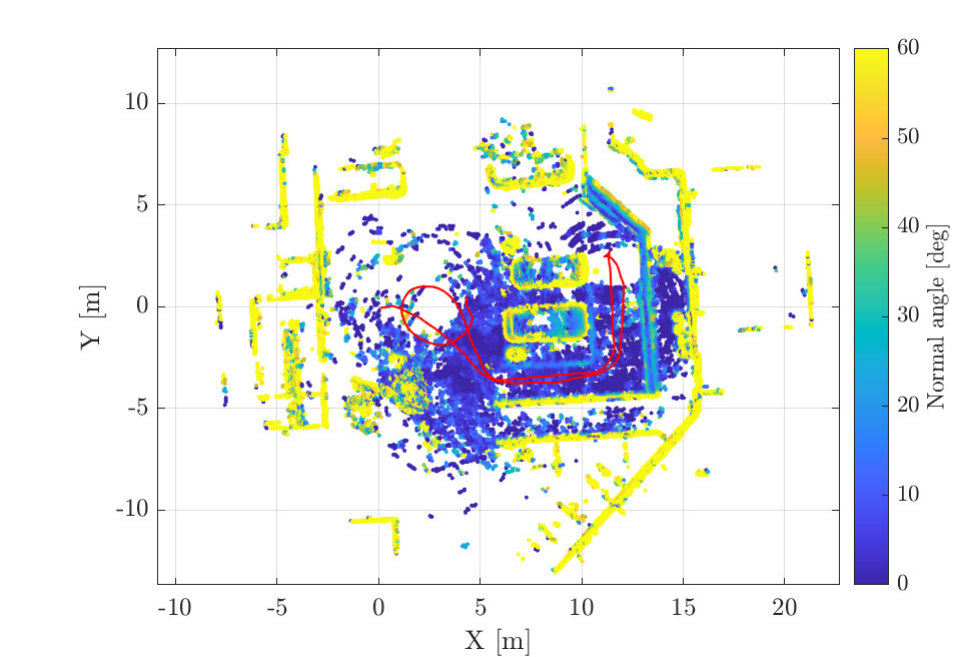}%
  \label{fig:building_9_obstacle_normals}}%
  \hfil%
  \subfloat[]{\includegraphics[width=0.249\textwidth]{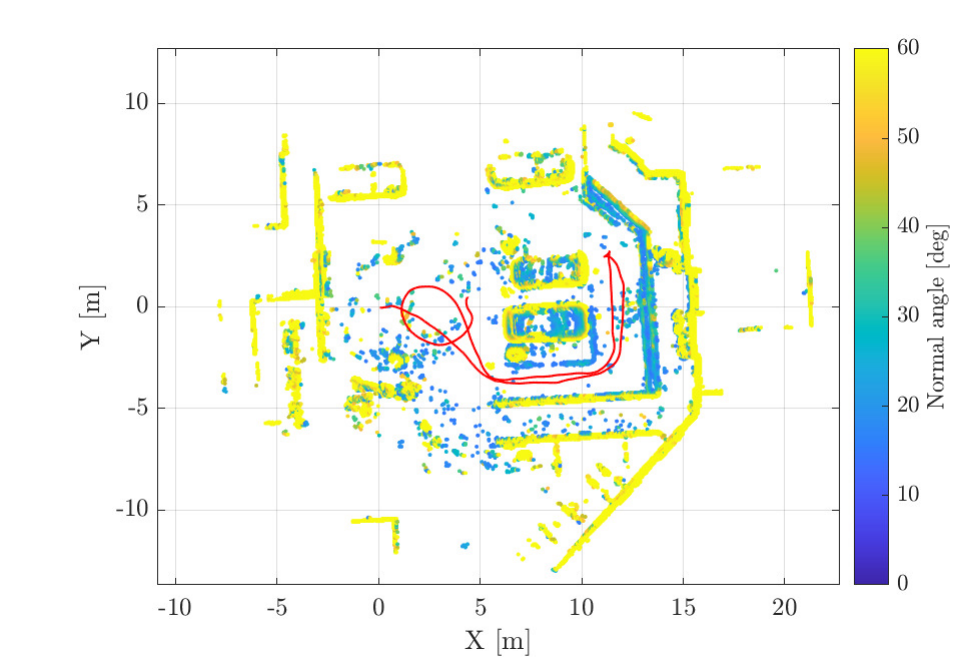}%
  \label{fig:building_9_obstacle_normals_vertical_height_filt}}%
  \hfil%
  \subfloat[]{\includegraphics[width=0.226\textwidth]{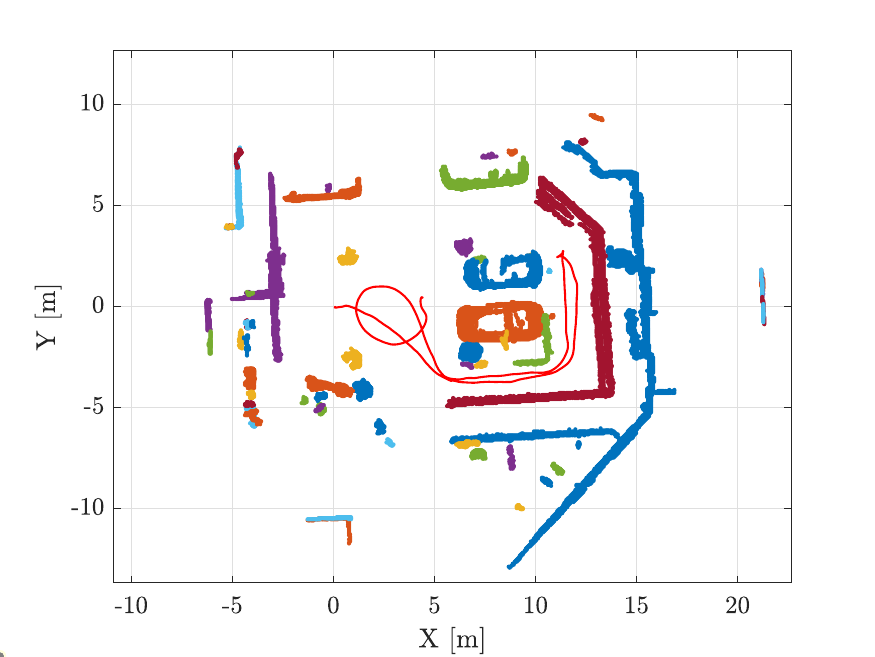}%
  \label{fig:building_9_obstacle_normals_vertical_clustered}}
  \hfil
  \subfloat[]{\includegraphics[width=0.226\textwidth]{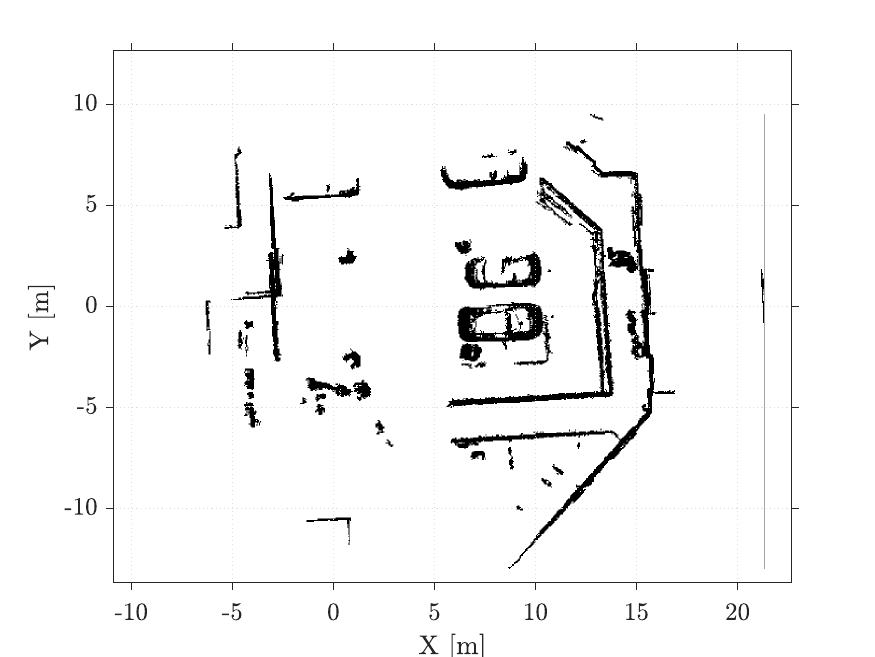}%
  \label{fig:building_9_obstacle_map}}
  \hfil%
  \caption{Intermediate steps of the positive obstacles pipeline. Centroids of the occupied cells color coded with the normal angle \protect\subref{fig:building_9_obstacle_normals}, points with normal far from the vertical after the height filter \protect\subref{fig:building_9_obstacle_normals_vertical_height_filt}, the points after the clustering step, that removed all small noisy clusters \protect\subref{fig:building_9_obstacle_normals_vertical_clustered} and the final 2D map with the cells marked by the detected points \protect\subref{fig:building_9_obstacle_map}. The red line represents the mapping trajectory.}
  \label{fig:building_9_obstacle}
\end{figure*}

Figure \ref{fig:building_9_octree} is a representation of the three-dimensional occupancy grid of the first map.
Notably, both the ground and other horizontal surfaces (the ceiling of the room towards the top of the image) are visible.
The accumulation of the point clouds achieves a reasonable point density over the mapped area, with the exception of a few locations like the triangular shadow behind the ramp in the top right area.
We expect those areas to be marked as not explored in the following steps.
Figure \ref{fig:building_9_ground} shows the different steps of the explored area pipeline.
The point cloud extracted from occupied cells of the occupancy grid is shown in Figure \ref{fig:building_9_ground_normals}, color coded with the normal angle $\alpha$.
Points with normal angle close to vertical are then clustered, as represented in Figure \ref{fig:building_9_ground_normals_flat_clustered}.
Notice that the ceiling to the right of the map belongs to a distinct cluster with respect to the one traversed by the trajectory, and for this reason gets discarded in the following steps.
Figures \ref{fig:building_9_ground_elevation} and \ref{fig:building_9_ground_map_inflated} show the two outputs of this block: the elevation map and explored area map, respectively.
As expected, the area behind the ramp is classified as unexplored.
Figure \ref{fig:building_9_obstacle} shows the intermediate steps of the positive obstacles pipeline.
As stated in Section \ref{subsec:positive_obstacles} the occupancy threshold for the extraction of occupied occupancy grid cells is much higher in this pipeline with respect to the explored area one to avoid false negatives.
This results in much less points being present in figure \ref{fig:building_9_obstacle_normals} with respect to \ref{fig:building_9_ground_normals}.
The points remaining after the elevation and normal angle filters are shown in Figure \ref{fig:building_9_obstacle_normals_vertical_height_filt}.
Most of these points clearly belong to building walls and other obstacles, but a relevant number of misclassified points is also visible.
This noise is removed by applying the Euclidean clustering algorithm and discarding small clusters (Figure \ref{fig:building_9_obstacle_normals_vertical_clustered}).
The final positive obstacles map is shown in Figure \ref{fig:building_9_obstacle_map}.

\begin{figure*}[!t]
  \centering
  \subfloat[]{\includegraphics[height=1.7in]{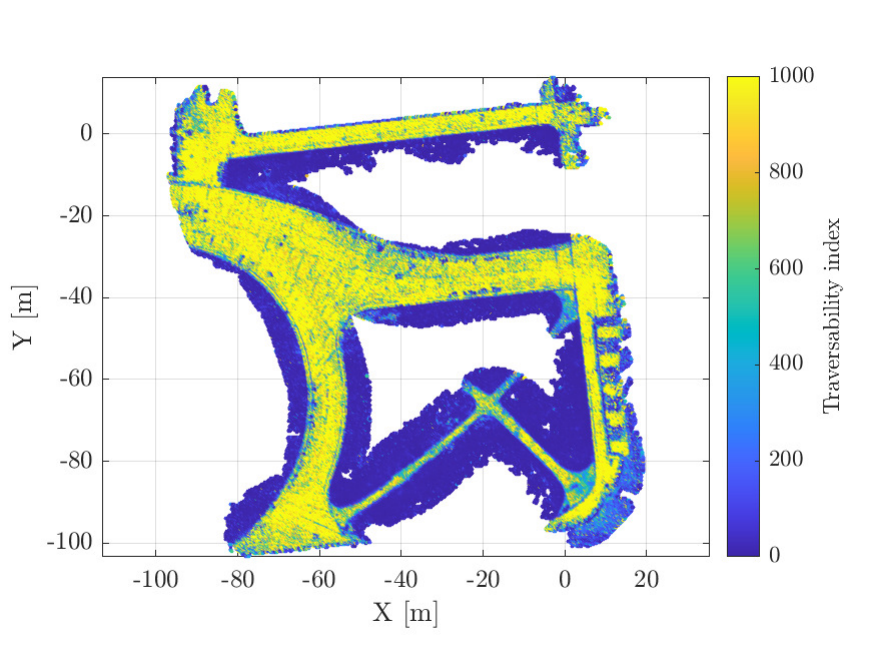}%
  \label{fig:piazza_leonardo_traversability_index}}
  \hfil%
  \subfloat[]{\includegraphics[height=1.7in]{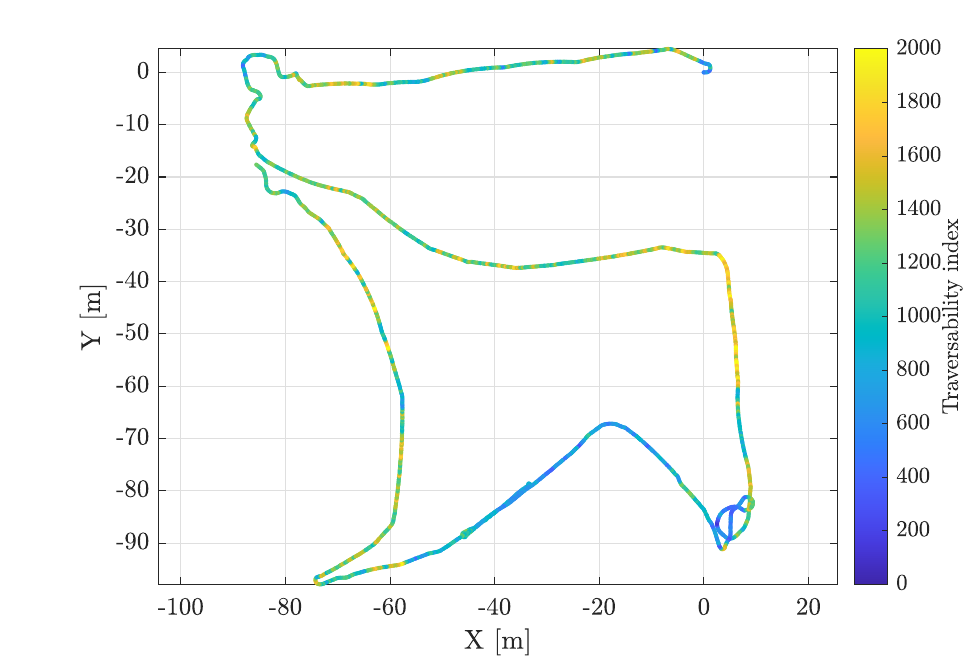}%
  \label{fig:piazza_leonardo_traversability_trajectory}}
  \hfil%
  \subfloat[]{\includegraphics[height=1.7in]{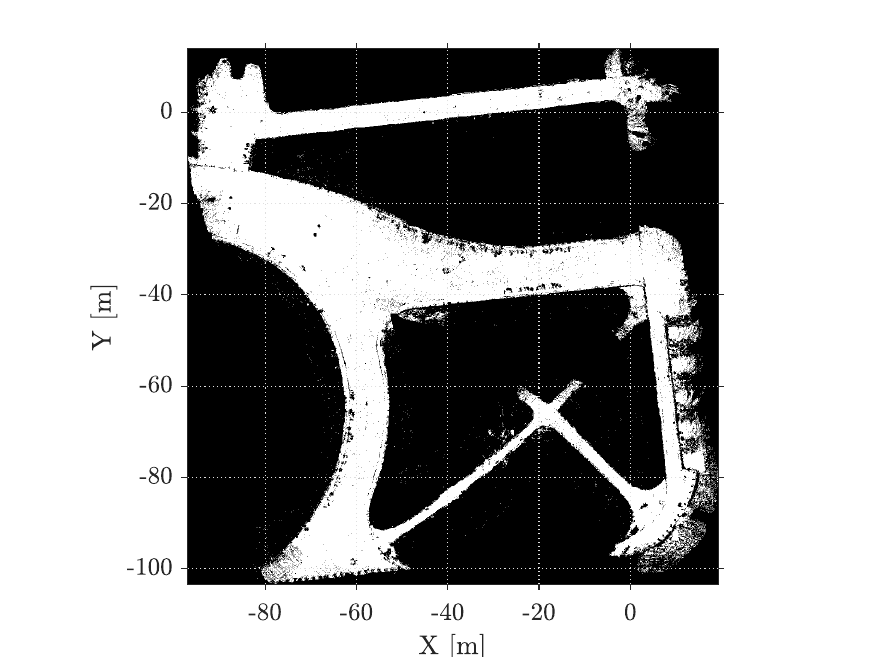}%
  \label{fig:piazza_leonardo_traversability_map}}
  \hfil%
  \caption{Intermediate steps of the traversability classification pipeline. The traversability index map obtained by merging all the classified point clouds \protect\subref{fig:piazza_leonardo_traversability_index}, the traversability values associated with the mapping trajectory \protect\subref{fig:piazza_leonardo_traversability_trajectory}}
  \label{fig:piazza_leonardo_trav}
\end{figure*}

The remaining modules are illustrated by exploiting the map of Piazza Leonardo da Vinci.
Figure \ref{fig:piazza_leonardo_traversability_index} shows the traversability index map, with the values associated with the mapping trajectory used as thresholds in Figure \ref{fig:piazza_leonardo_traversability_trajectory}.
The resulting thresholded map in Figure \ref{fig:piazza_leonardo_traversability_map} clearly marks the edges between different surfaces.

\begin{figure*}[!t]
  \centering
  \subfloat[]{\includegraphics[width=0.249\textwidth]{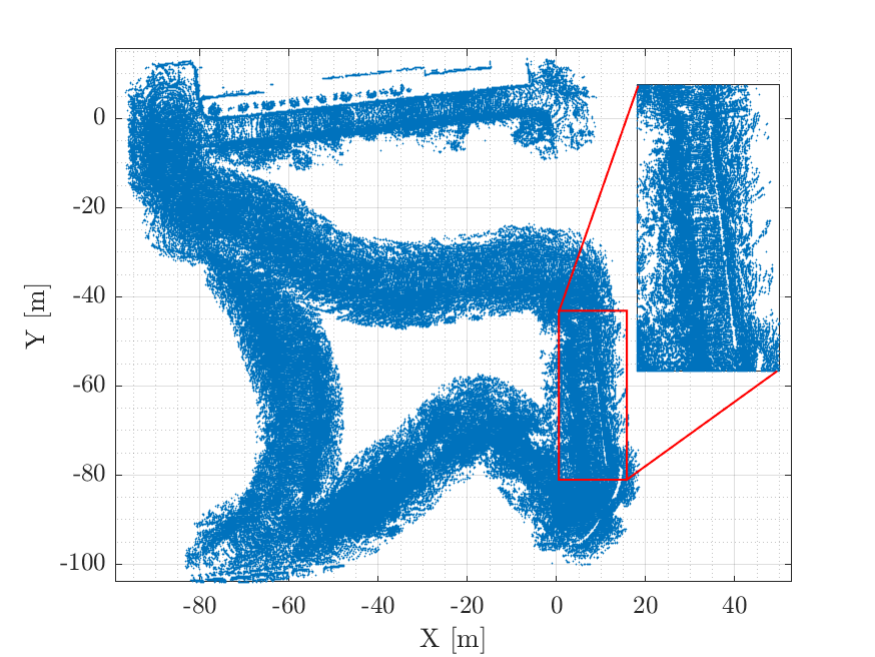}%
  \label{fig:piazza_leonardo_negative_obstacles_not_expanded}}
  \hfil%
  \subfloat[]{\includegraphics[width=0.249\textwidth]{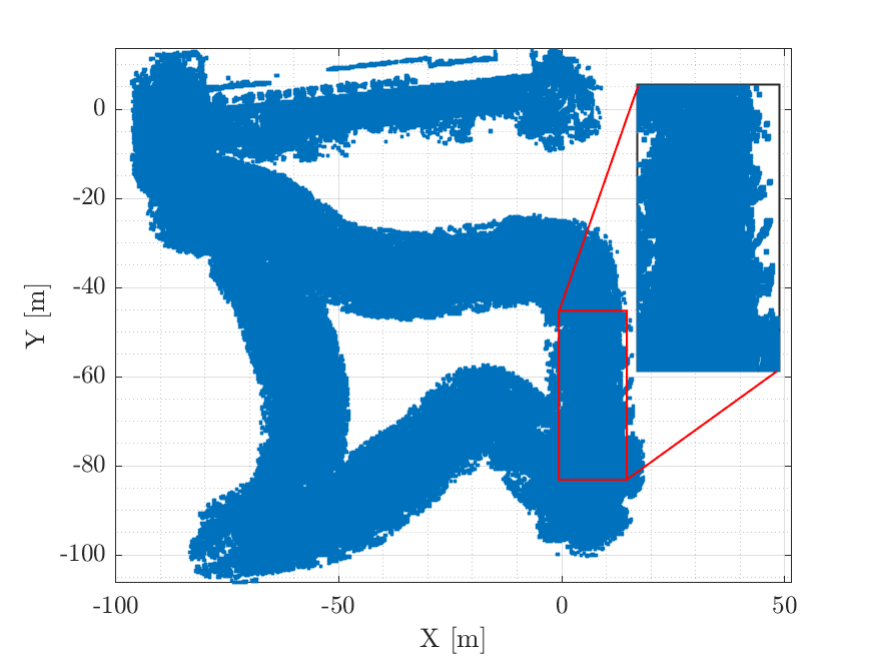}%
  \label{fig:piazza_leonardo_negative_obstacles_expanded}}
  \hfil%
  \subfloat[]{\includegraphics[width=0.249\textwidth]{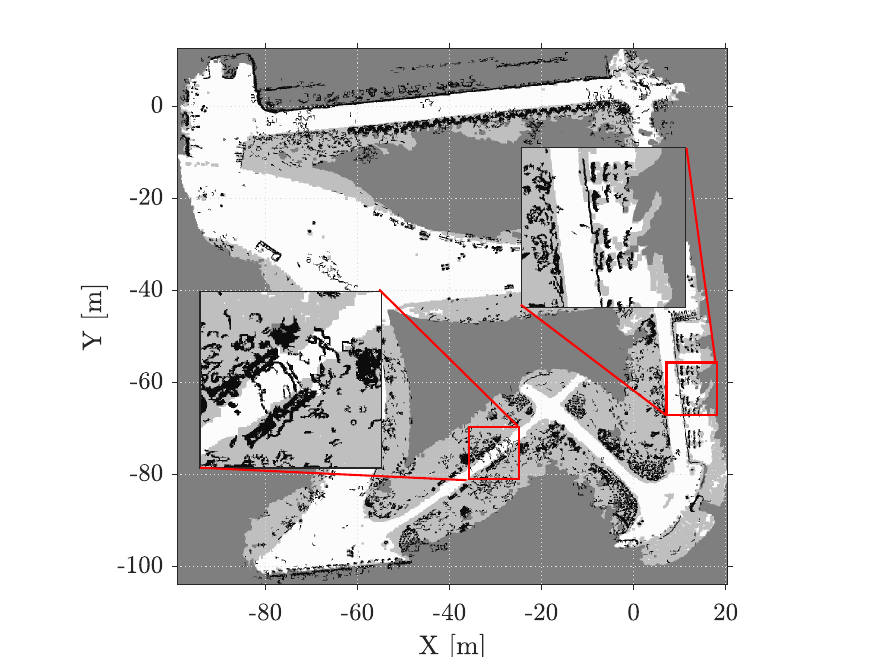}%
  \label{fig:piazza_leonardo_negative_obstacles_unfiltered_zoom}}
  \hfil%
  \subfloat[]{\includegraphics[width=0.249\textwidth]{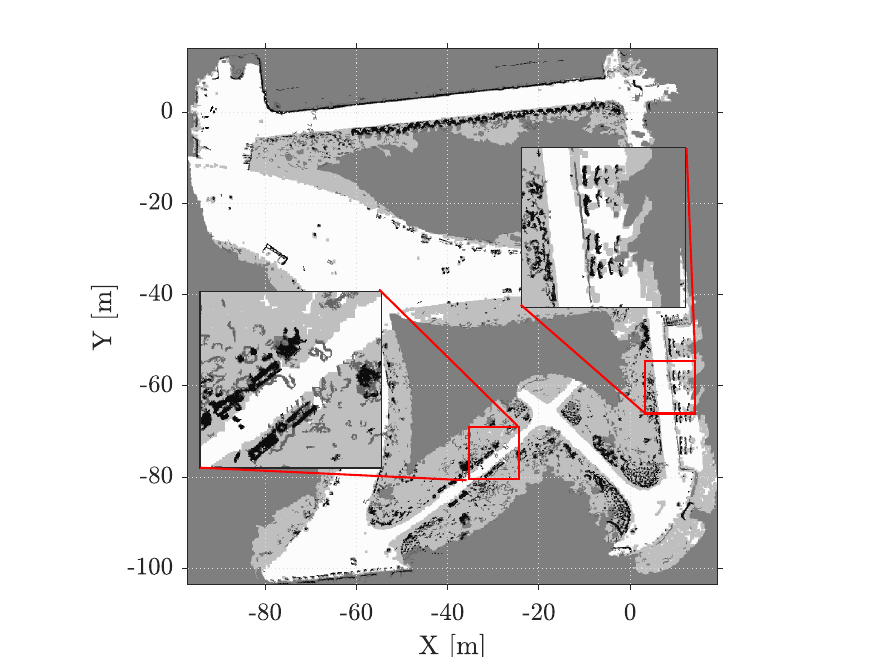}%
  \label{fig:piazza_leonardo_negative_obstacles_filtered_zoom}}
  \hfil%
  \caption{Top view of the 3D map before \protect\subref{fig:piazza_leonardo_negative_obstacles_not_expanded} and after \protect\subref{fig:piazza_leonardo_negative_obstacles_expanded} the expansion in the negative obstacles pipeline. Notice the lack of points in correspondence with the negative obstacle in \protect\subref{fig:piazza_leonardo_negative_obstacles_not_expanded}. Unfiltered obstacles computed after the expansion of the map correctly detect the negative obstacle but also cause false negatives \protect\subref{fig:piazza_leonardo_negative_obstacles_unfiltered_zoom}. After the filtering \protect\subref{fig:piazza_leonardo_negative_obstacles_filtered_zoom} the effect of false positives is mitigated.}
  \label{fig:piazza_leonardo_neg_obs}
\end{figure*}

Finally, Figure \ref{fig:piazza_leonardo_neg_obs} illustrates some details about the negative obstacles pipeline.
Figure \ref{fig:piazza_leonardo_negative_obstacles_not_expanded} and \ref{fig:piazza_leonardo_negative_obstacles_expanded} compare the 3D map before and after the iterative expansion, respectively.
Notice how before the expansion no points can be found close to the negative obstacle.
Figure \ref{fig:piazza_leonardo_negative_obstacles_unfiltered_zoom} and \ref{fig:piazza_leonardo_negative_obstacles_filtered_zoom} compare the extracted obstacles with and without the filtering step exploiting the traversable area.
\subsection{Quantitative validation}
\label{subsec:quantitative_validation}
\begin{figure*}[!t]
  \centering
  \subfloat[]{\includegraphics[width=0.48\textwidth]{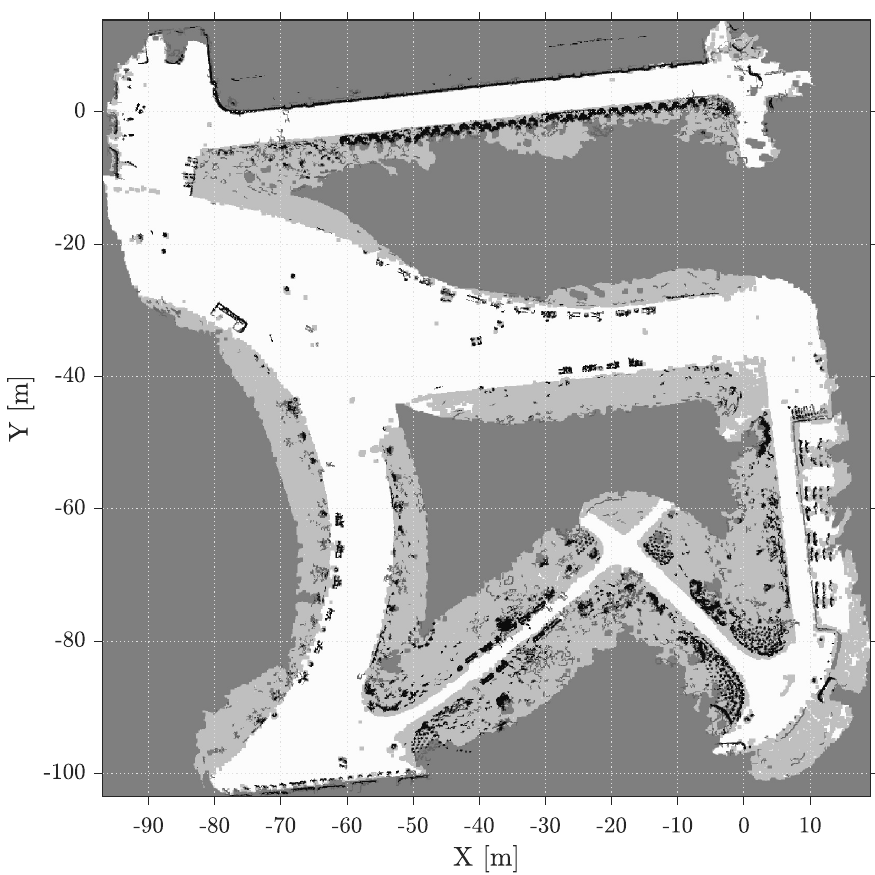}%
  \label{fig:piazza_leonardo_final}}
  \hfil%
  \subfloat[]{\includegraphics[width=0.48\textwidth]{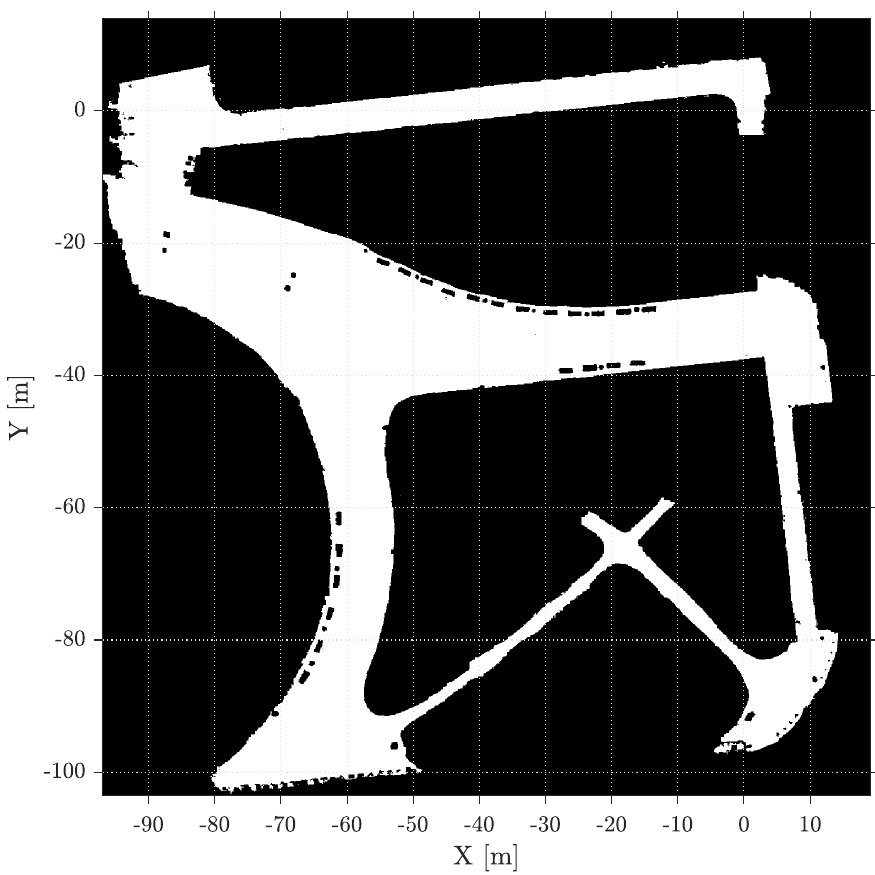}%
  \label{fig:piazza_leonardo_truth}}
  \caption{Final navigation map with all the four layers \protect\subref{fig:piazza_leonardo_final} and the ground truth map \protect\subref{fig:piazza_leonardo_truth}}
\end{figure*}
\begin{figure}[!t]
  \centering
  \includegraphics[height=1.4in]{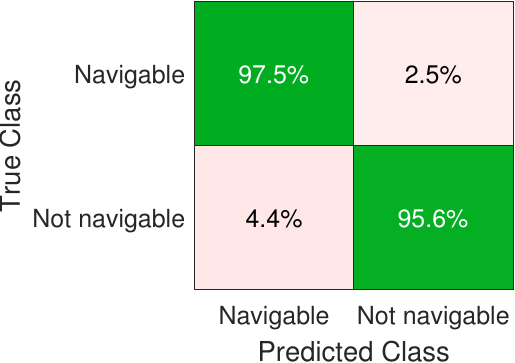}
  \caption{Classification metrics: pixel-by-pixel}
  \label{fig:classification_metrics}
\end{figure}

To validate the mapping algorithm, the Piazza Leonardo map was chosen, as it is a realistic navigation scenario and contains all the different features of interest.
A ground truth map was created by hand labelling the pixels, as shown in figure \ref{fig:piazza_leonardo_truth}.
Classification metrics were then obtained by comparing the two maps pixel by pixel.
Note that in map \ref{fig:piazza_leonardo_final} a pixel is considered to be traversable only if it is white (that is, it is considered traversable by all layers).
Classification metrics are reported in Figure \ref{fig:classification_metrics}, which shows excellent performances.

\begin{figure}[!t]
  \centering
  \includegraphics[height=1.4in]{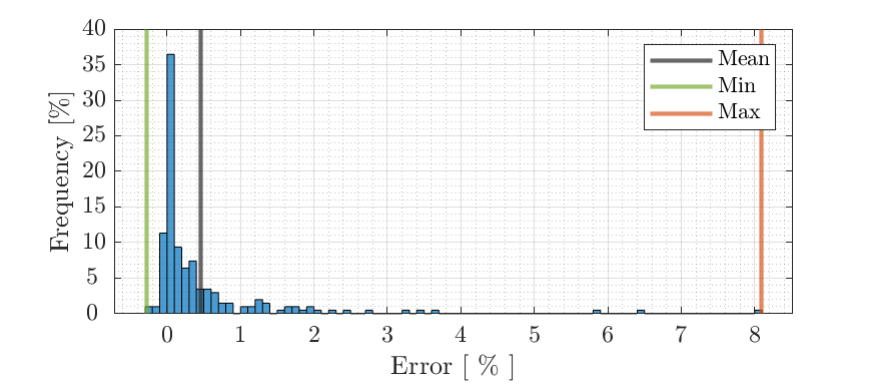}
  \caption{Distribution of path length errors in the Monte Carlo validation}
  \label{fig:piazza_leonardo_monte_carlo_histogram}
\end{figure}

In the context of motion planning, however, pure classification performance is not indicative of a good quality map.
A small number of false negatives could make the robot discard all trajectories through a path, forcing a long detour or making a region unreachable altogether.
Similarly, a few false positives could make the robot plan a trajectory through an obstacle, guiding it to a suboptimal or impossible path.
To evaluate the obtained map in a more realistic way a Monte Carlo approach was used.
First, the two maps were inflated, to take into account the robot's footprint.
Then $n=1000$ start-goal tuples were randomly extracted.
Note that the starting position is always extracted within the navigable region of the ground truth map, since it represents the physical location of the robot, which cannot be in non-navigable space.
For each tuple the $A^*$ planning algorithm was applied on both maps, storing the results.
On the ground truth map successful planning was achieved in $220$ occurrences, of which $203$ ($92.30 \%$) were also successful on the autogenerated map.
Among the $17$ failures the vast majority ($16$) was due to the misclassification of the goal position, while the remaining one was caused by map connectivity issues.
Complementarily, on the ground truth map $780$ planning requests failed, of which $772$ ($99.00 \%$) were also rejected on the auto-generated map.
All the planning requests that were wrongly successful were due to the misclassification of the goal point, no connectivity issues were identified.
A further analysis can be conducted on the true positive cases: to highlight any connectivity issue forcing a long re-routing, for each path the difference in length on the two maps is computed.
Figure \ref{fig:piazza_leonardo_monte_carlo_histogram} shows the more than satisfactory results of this metric, with an average overlength of just $0.33 m$
\begin{figure}[!t]
  \centering
  \includegraphics[height=1.4in]{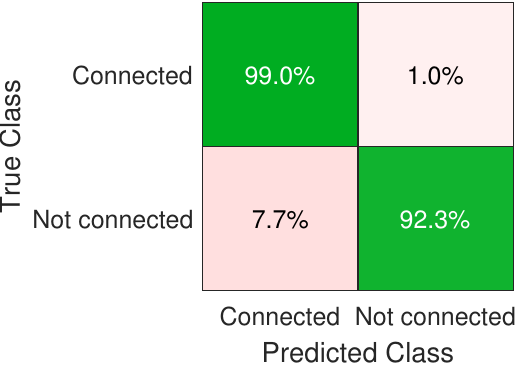}
  \caption{Classification metrics: Monte Carlo simulations}
  \label{fig:classification_metrics_monte_carlo}
\end{figure}

\section{Conclusions}
The paper proposes an algorithm for autonomously generating navigation maps from LiDAR data.
The robot is first required to explore the environment while acquiring range measurements.
A SLAM algorithm is then used to obtain a consistent localization and subsequently aggregate range measurements in a three-dimensional occupancy grid.
The occupancy grid then feeds the different modules of the algorithm, each dedicated to a specific obstacle type: unexplored areas, positive and negative obstacles, untraversable terrain.
Each module provides a binary navigability map, which are then fused together to obtain the full environment representation.
An experimental validation has been performed, which has shown excellent performances both in term of pixel-by-pixel classification and connectedness of the map.
\bibliography{IEEEabrv,bibliography}
\bibliographystyle{IEEEtran}
\begin{IEEEbiography}[{\includegraphics[width=1in,clip,keepaspectratio]{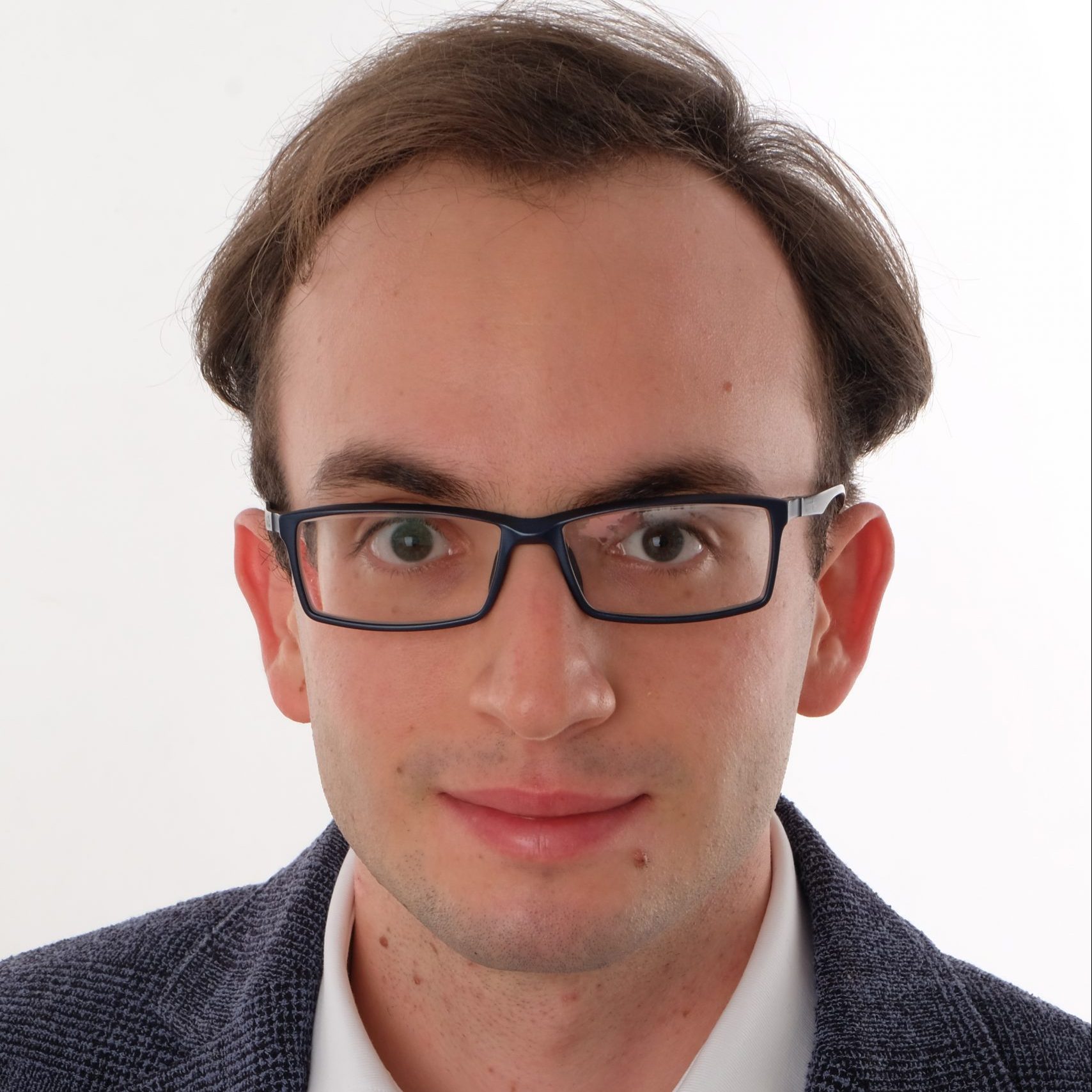}}]{Luca Mozzarelli}%
  was born near Milan, Italy, in 1996. He received the Bachelor Degree on July 2018 and the Master of Science cum laude on December 2020, both in Automation and Control Engineering from Politecnico di Milano (Italy).
  He is currently working towards the PhD degree in Information Engineering with the Dipartimento di Elettronica, Informazione e Bioingegneria.
  His research interests include mapping, localization and planning algorithms for mobile robots and off-road vehicles.
\end{IEEEbiography}
\begin{IEEEbiography}[{\includegraphics[width=1in,height=1.25in,clip,keepaspectratio]{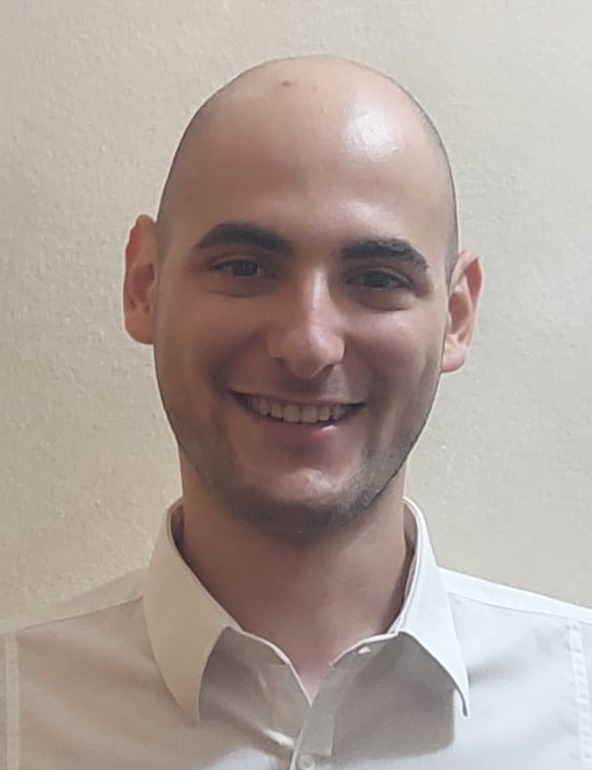}}]{Simone Specchia}
  was born near Lecce, Italy, in 1997. He received the Bachelor Degree on July 2019 and the Master of Science cum laude on December 2021, in Automation and Control Engineering from Politecnico di Milano (Italy).
  He is currently working towards the PhD degree in Information Engineering with the Dipartimento di Elettronica, Informazione e Bioingegneria.
  His research interests include autonomous navigation and dynamics and control of vehicles.
\end{IEEEbiography}
\begin{IEEEbiography}[{\includegraphics[width=1in,height=1.25in,clip,keepaspectratio]{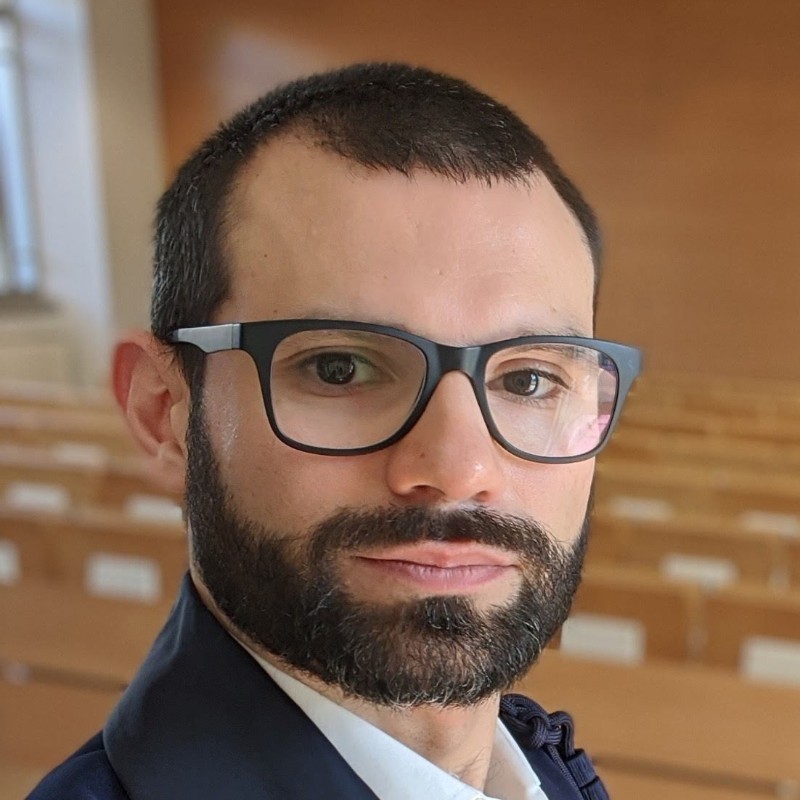}}]{Matteo Corno}
  received the M.Sc. degree in computer and electrical engineering from the University of Illinois, Champaign, IL, USA, in 2005, and the Ph.D. degree in Information engineering (system control specialization) from the Politecnico di Milano, Milan, Italy, in 2009.
  He is currently an Associate Professor with the Dipartimento di Elettronica, Informazione e Bioingegneria, Politecnico di Milano.
  He held research positions with the University of Minnesota, Minneapolis, MN, USA, Johannes Kepler University, Linz, Austria, and TU Delft, Delft, The Netherlands.
  His research interests include dynamics and control of vehicles, and lithium-ion battery modeling, estimation, and control.
\end{IEEEbiography}
\begin{IEEEbiography}[{\includegraphics[width=1in,height=1.25in,clip,keepaspectratio]{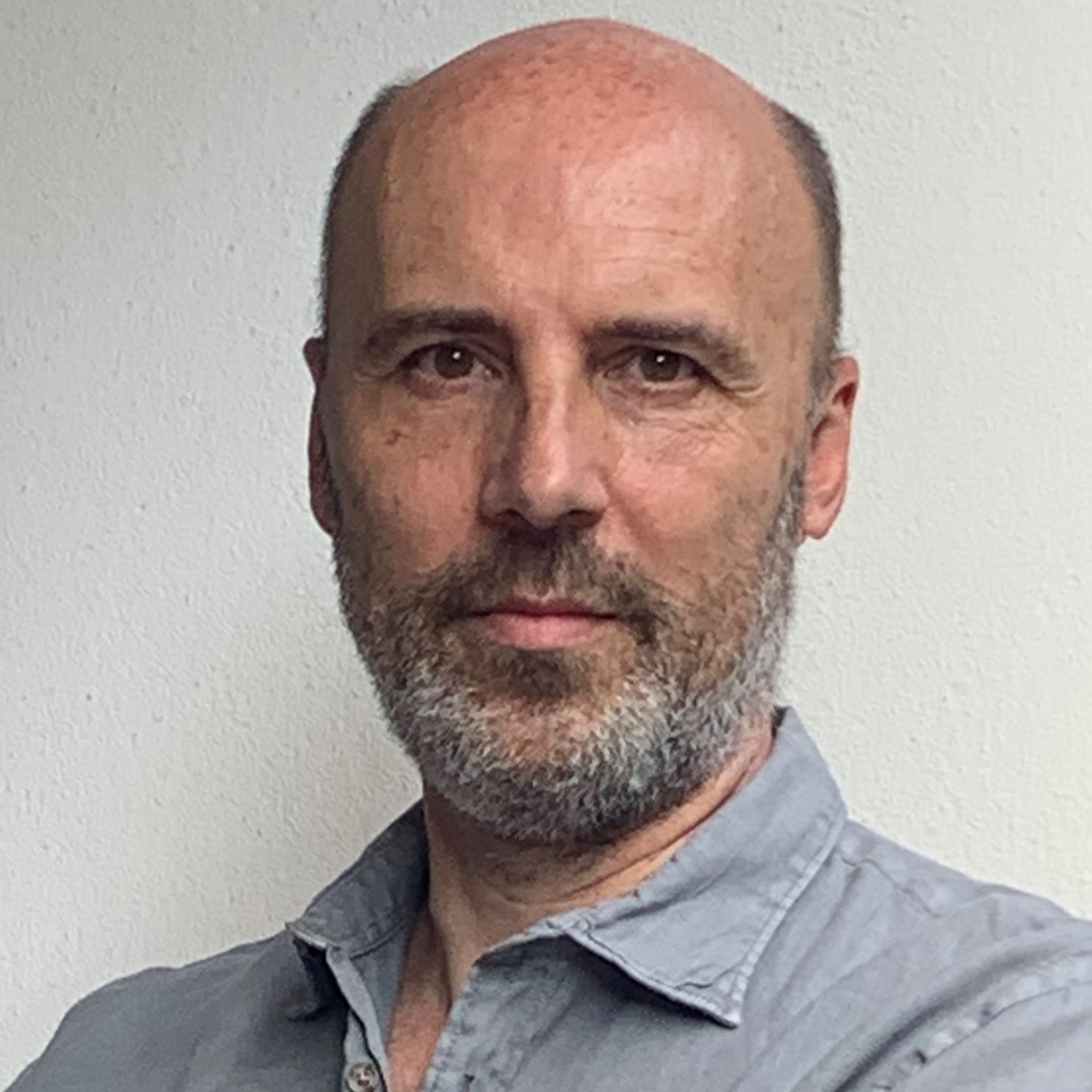}}]{Sergio Matteo Savaresi}
  received the M.Sc. in ElectricalEngineering (Politecnico di Milano, 1992), the Ph.D. in Systems and Control Engineering (Politecnico di Milano, 1996), and the M.Sc. in AppliedMathematics (Catholic University, Brescia, 2000).
  After the Ph.D. he worked as management consultant at McKinsey\&Co, Milan Office.
   He is FullProfessor in Automatic Control at Politecnico di Milano since 2006.
   He is Deputy Director and Chair of the Systems\&Control Section of Department of Electronics, Computer Sciences and Bioengineering (DEIB), Politecnico di Milano.
   He is author of more than 500 scientific publications.
   His main interests are in the areas of vehicles control, automotive systems, data analysis and system identification, non-linear control theory,and control applications, with special focus on smart mobility.
   He has been manager and technical leader of more than 400 research projects in cooperation with private companies.
   He is co-founder of 8 high-tech startup companies.
\end{IEEEbiography}

\end{document}